\documentclass[10pt,twocolumn,letterpaper]{article}
\usepackage[pagenumbers]{cvpr}
\usepackage{cuted} %

\usepackage{currfile} %

\usepackage{caption} %
\usepackage{amssymb}
\usepackage{pifont}
\usepackage{colortbl} %

\newcommand{\xmark}{\ding{55}}

\definecolor{cvprblue}{rgb}{0.21,0.49,0.74}
\usepackage[breaklinks,colorlinks,allcolors=cvprblue]{hyperref}
\hypersetup{
  pdftitle={HighwayVerse: A Chain-of-Inspection Benchmark for Highway Infrastructure Anomaly Understanding},
  pdfauthor={Junyi Yang, Zheng Si, Yilin Zhai, Pengpeng Ouyang, Hongxi Wang, Kangze Li, Mo Chen, Sicong Zhang, Hua Wang, Dongyuan Wu, Dong Chen, Mingliang Xu},
  pdfsubject={Highway infrastructure inspection benchmark},
  pdfkeywords={HighwayVerse, LaneRAG, visual language models, highway inspection}
}
\let\citesupp\cite
\title{HighwayVerse: A Chain-of-Inspection Benchmark for Highway Infrastructure Anomaly Understanding}
\author{%
Junyi Yang\textsuperscript{1}, Zheng Si\textsuperscript{1}, Yilin Zhai\textsuperscript{1}, Pengpeng Ouyang\textsuperscript{1}, Hongxi Wang\textsuperscript{1}, Kangze Li\textsuperscript{1}\\[2pt]
Mo Chen\textsuperscript{1}, Sicong Zhang\textsuperscript{2}, Hua Wang\textsuperscript{1}, Dongyuan Wu\textsuperscript{1}, Dong Chen\textsuperscript{1,}\thanks{Corresponding Author.}, Mingliang Xu\textsuperscript{1}\\[7pt]
{\normalsize \textsuperscript{1}Zhengzhou University\qquad \textsuperscript{2}Monash University}
}

\begin{document}
\maketitle
\raggedbottom
\begin{abstract}
Practical highway infrastructure inspection entails progressive diagnostic reasoning from visual perception to maintenance decisions, yet existing benchmarks remain confined to isolated tasks, leaving Multimodal Large Language Models (MLLMs) insufficiently evaluated across sequential inspection stages. Meanwhile, current datasets focus primarily on highway pavement defects, falling short of real-world inspection across diverse infrastructure components such as highway pavements, guardrails, glare screens, and drainage systems. To address these gaps, we formulate the \textbf{Highway Infrastructure Inspection Progression Chain (HIIP-Chain)}, spanning \textbf{Anomaly Detection}, \textbf{Anomaly Recognition}, \textbf{Condition Analysis}, and \textbf{Maintenance Resolution}, and introduce \textbf{HighwayVerse}. HighwayVerse comprises 14,143 real-world Ground and Unmanned Aerial Vehicle (UAV) inspection records spanning 100 and 17 anomaly categories, respectively, with 84,858 expert-reviewed question-answer pairs. We evaluate 12 open-source and closed-source MLLMs using the chained metric \textbf{Final}, which credits a correct maintenance decision only when all preceding diagnostic stages are completed correctly. Results show that although models achieve relatively high accuracy when initially judging whether an anomaly is present, their markedly lower \textbf{Final} scores expose struggles across downstream recognition, analysis, and decision-making. This discrepancy indicates that current MLLMs do not fully comprehend specific highway anomaly patterns beyond superficial initial judgments. Furthermore, we propose \textbf{\mbox{LaneRAG}}, a training-free retrieval-augmented framework that provides hierarchical reference guidance achieving consistent performance improvements across diverse MLLMs. Our dataset, evaluation benchmark, and source code will be publicly released.
\end{abstract}

\section{Introduction}
\label{sec:intro}

Highway infrastructure, including highway pavements, guardrails, glare screens, and drainage systems, is continuously exposed to heavy traffic loads and harsh environmental conditions. Timely and accurate anomaly inspection is essential for maintaining structural integrity, ensuring transportation safety, and reducing maintenance costs~\cite{Spencer_2019_InfrastructureInspection,Zhu_2022_UAVPavement}.

\begin{figure*}[t]
\begingroup
\centering
\captionsetup{skip=4pt}
\captionsetup[subfigure]{skip=3pt}
\begin{minipage}{0.99\textwidth}
\centering
\begin{subfigure}[b]{0.40\linewidth}
\centering
\begin{minipage}{0.84\linewidth}
\centering
{\normalfont\rmfamily\scriptsize\bfseries Ground}\par\vspace{1pt}
\includegraphics[width=\linewidth]{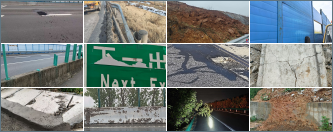}\par
\vspace{3pt}
{\normalfont\rmfamily\scriptsize\bfseries UAV}\par\vspace{1pt}
\includegraphics[width=\linewidth]{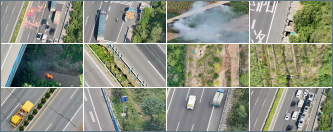}
\end{minipage}
\caption{Ground and UAV examples.}
\label{fig:highwayverse_examples}
\end{subfigure}\hfill
\begin{subfigure}[b]{0.58\linewidth}
\centering
\includegraphics[width=\linewidth,height=54mm,keepaspectratio]{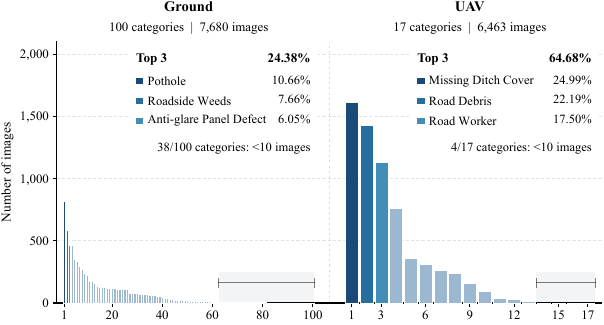}
\caption{Category distributions across viewpoints.}
\label{fig:category_distribution}
\end{subfigure}

\vspace{1pt}
\begin{subfigure}[t]{\linewidth}
\centering
\includegraphics[width=\linewidth]{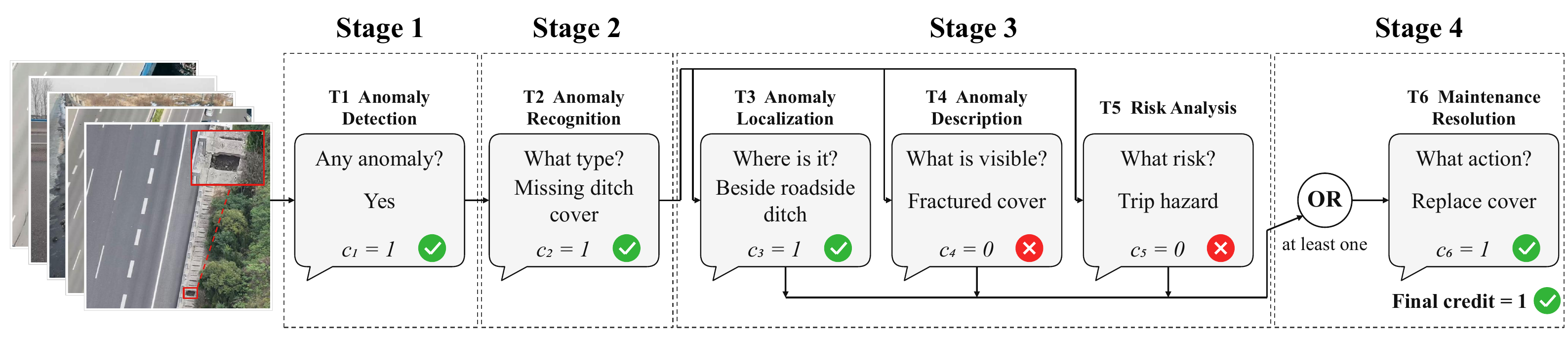}
\caption{Six inspection questions across four stages with dependency-aware scoring.}
\label{fig:highwayverse_overview_c}
\end{subfigure}
\end{minipage}
\caption{Overview of HighwayVerse. (a) Twelve anomalous examples per viewpoint from independent, unpaired sources. (b) Ranked category frequencies for Ground and UAV, with top-three and tail statistics. (c) Six questions span four stages. Final credit requires correct Anomaly Detection, Anomaly Recognition, Maintenance Resolution, and at least one correct Condition Analysis answer.}
\label{fig:highwayverse_overview}
\endgroup
\end{figure*}

Conventional automated inspection primarily relies on task-specific detectors and segmenters~\cite{Maeda_2018_RoadDamage,Zhou_2024_CrackSeg}, which focus on localized visual perception. In practice, highway infrastructure inspection involves a progressive diagnostic process in which inspectors verify anomalies, recognize and localize them, characterize their conditions, evaluate associated risks, and determine appropriate maintenance actions~\cite{Spencer_2019_InfrastructureInspection}. We formalize this progression of interdependent diagnostic judgments as the \textbf{Highway Infrastructure Inspection Progression Chain (HIIP-Chain)}, a structured framework linking visual anomaly understanding to maintenance-oriented decisions. Toward this goal, Multimodal Large Language Models (MLLMs) offer strong potential by integrating visual perception with domain-contextualized reasoning across all stages~\cite{Bai_2025_Qwen3VL,Wang_2025_InternVL35}. However, existing benchmarks still fall short of this complete progression, where RoadBench covers road damage classification and defect description~\cite{Xiao_2026_RoadBench}, while DriveLM, RoadSocial, and RoadSceneBench focus on object perception, traffic event analysis, motion prediction, and driving action planning~\cite{Sima_2024_DriveLM,Parikh_2025_RoadSocial,Liu_2026_RoadSceneBench}. These benchmarks do not assess maintenance correctness under the diagnostic prerequisites of the HIIP-Chain, leaving a critical gap in highway infrastructure inspection evaluation.

A further limitation lies in the restricted variety of infrastructure entities and anomaly categories covered by current datasets. Ground-level datasets, which capture fine-grained anomalies at close range, such as CrackSeg~\cite{Zhou_2024_CrackSeg} and RoadAnomaly~\cite{Lis_2019_RoadAnomaly}, focus on pavement cracks and unexpected roadway obstacles. Meanwhile, Unmanned Aerial Vehicle (UAV) imagery covers long road segments efficiently, yet datasets incorporating this perspective, such as RDD2022~\cite{Arya_2024_RDD2022} and UAPD~\cite{Zhu_2022_UAVPavement}, remain restricted to pavement distress, with RDD2022 covering only four damage types. In contrast, real-world highway infrastructure spans far more diverse entities, giving rise to a much broader and more complex spectrum of anomaly categories. Existing datasets therefore fall short of real-world inspection demands and cannot adequately evaluate MLLMs on practical inspection tasks.

To address these limitations, we introduce \textbf{HighwayVerse}, a multi-view benchmark designed to evaluate MLLMs across the complete HIIP-Chain. In terms of scale, HighwayVerse comprises 14,143 real-world inspection records, including 7,680 Ground records across 100 anomaly categories and 6,463 UAV records across 17 categories, as shown in Figure~\ref{fig:highwayverse_overview}\subref{fig:highwayverse_examples}. Furthermore, these records naturally preserve the long-tailed category distributions of field inspection depicted in Figure~\ref{fig:highwayverse_overview}\subref{fig:category_distribution}. Specifically, each record contains six multiple-choice questions organized into four progressive stages, namely \textbf{Anomaly Detection}, \textbf{Anomaly Recognition}, \textbf{Condition Analysis} covering parallel \textbf{Anomaly Localization}, \textbf{Anomaly Description}, and \textbf{Risk Analysis}, and \textbf{Maintenance Resolution}, as illustrated in Figure~\ref{fig:highwayverse_overview}\subref{fig:highwayverse_overview_c}. Crucially, across these four stages, a correct answer in a later stage receives credit only when the preceding stages are completed correctly. In total, the dataset contains 84,858 question-answer pairs generated semi-automatically and reviewed by domain-trained annotators over 400 hours.

\begin{table*}[!t]
  \centering
  \captionsetup{position=top,font=small}
  \caption{Comparison with road-related datasets and benchmarks. Counts exclude background; \checkmark, $\triangle$, and \xmark{} denote explicit, limited, and no reported coverage, and -- denotes unreported or inapplicable splits. Infra.\ defects and Maint.\ denote infrastructure defect and maintenance tasks, while Inspection chain links Anomaly Detection, Anomaly Recognition, Condition Analysis, and Maintenance Resolution.}
  \label{tab:comparison}
  \small
  \setlength{\tabcolsep}{3.2pt}
  \renewcommand{\arraystretch}{1.12}
  \begin{tabular*}{\linewidth}{@{\extracolsep{\fill}}lccccccccc@{}}
      \toprule
      & & & \multicolumn{2}{c}{\textbf{Viewpoints}} & \multicolumn{5}{c}{\textbf{Task coverage}} \\
      \cmidrule(lr){4-5}\cmidrule(lr){6-10}
      \shortstack{\textbf{Dataset /}\\\textbf{benchmark}} & \textbf{Domain} & \shortstack{\textbf{Anomaly}\\\textbf{categories}} & \textbf{Ground} & \textbf{UAV} & \shortstack{\textbf{Infra.}\\\textbf{defects}} & \textbf{QA} & \textbf{Reasoning} & \textbf{Maint.} & \shortstack{\textbf{Inspection}\\\textbf{chain}} \\
      \midrule
      RoadAnomaly~\cite{Lis_2019_RoadAnomaly} & Road & -- & \checkmark & \xmark & \xmark & \xmark & \xmark & \xmark & \xmark \\
      CrackSeg~\cite{Zhou_2024_CrackSeg} & Pavement & 1 & \checkmark & \xmark & \checkmark & \xmark & \xmark & \xmark & \xmark \\
      RDD2022~\cite{Arya_2024_RDD2022} & Pavement & 4 & \checkmark & \checkmark & \checkmark & \xmark & \xmark & \xmark & \xmark \\
      UAPD~\cite{Zhu_2022_UAVPavement} & Pavement & 6 & \xmark & \checkmark & \checkmark & \xmark & \xmark & \xmark & \xmark \\
      RoadBench\textsuperscript{\dag}~\cite{Xiao_2026_RoadBench} & Pavement & 10 & -- & -- & \checkmark & \checkmark & \xmark & \xmark & \xmark \\
      DriveLM~\cite{Sima_2024_DriveLM} & Driving & -- & \checkmark & \xmark & \xmark & \checkmark & \checkmark & \xmark & \xmark \\
      RoadSocial~\cite{Parikh_2025_RoadSocial} & Road & -- & \checkmark & \checkmark & $\triangle$ & \checkmark & \checkmark & \xmark & \xmark \\
      RoadSceneBench~\cite{Liu_2026_RoadSceneBench} & Road & -- & \checkmark & \xmark & \xmark & \checkmark & \checkmark & \xmark & \xmark \\
      \midrule
      \textbf{HighwayVerse} & \textbf{Highway} & \textbf{100+17} & \checkmark & \checkmark & \checkmark & \checkmark & \checkmark & \checkmark & \checkmark \\
      \bottomrule
  \end{tabular*}
\end{table*}

Building upon HighwayVerse, we evaluate 12 representative open-source and closed-source MLLMs across Ground and UAV perspectives via \textbf{Final}, the overall chained accuracy requiring correct completion across all stages of the HIIP-Chain. Although evaluated models achieve high accuracy when initially determining whether an anomaly is present, their overall chained accuracy in Final drops substantially. On the UAV perspective, for instance, InternVL3.5-8B attains 72.48\% when initially judging whether an anomaly exists yet reaches only 25.96\% in Final. Because Final incorporates downstream recognition, condition analysis, and maintenance decisions, this sharp drop reveals widespread errors in anomaly-related analysis and decision-making, indicating that MLLMs likely fail to truly comprehend specific anomaly patterns and that their high initial scores largely stem from chance guessing. Compared with prior isolated task evaluations, our chain-structured HIIP-Chain protocol is therefore far more effective at assessing genuine MLLM understanding of highway infrastructure anomalies.

To further strengthen highway infrastructure inspection along the HIIP-Chain, we introduce Local-and-Global Navigated Exemplar Retrieval-Augmented Guidance (\textbf{\mbox{LaneRAG}}), a training-free adaptation framework for frozen MLLMs. Built upon an offline hierarchical reference bank that organizes highway infrastructure anomalies from coarse-grained to fine-grained categories, \mbox{LaneRAG} performs route-guided complementary retrieval across Primary, Expanded, and Global Scopes, assembling retrieved reference images with their category labels into a reference-conditioned prompt for in-context visual guidance. Empirically, pairing \mbox{LaneRAG} with all 12 evaluated MLLMs yields consistent and substantial improvements across the chain. On the UAV perspective, \mbox{LaneRAG} raises Anomaly Recognition and Final by 12.35\% and 6.75\% on the open-source InternVL3.5-8B, and by 7.99\% and 4.68\% on the closed-source Gemini 2.5 Flash.

Our main contributions are summarized below.
\begin{itemize}[leftmargin=*,topsep=2pt,itemsep=1pt,parsep=0pt,partopsep=0pt]
\item We construct \textbf{HighwayVerse}, a multi-view highway infrastructure dataset of 14,143 real-world inspection records and 84,858 expert-reviewed question-answer pairs, spanning 7,680 Ground records across 100 anomaly categories and 6,463 UAV records across 17 categories.
\item We formulate the \textbf{HIIP-Chain}, a progressive evaluation protocol aligned with engineering inspection workflows, providing a faithful standard to assess genuine MLLM comprehension of highway infrastructure anomalies across 12 open-source and closed-source models.
\item We propose \textbf{\mbox{LaneRAG}}, a companion retrieval-augmented framework that integrates reference images into a reference-conditioned prompt to substantially boost inspection performance across all evaluated MLLMs without updating any MLLM parameters.
\end{itemize}
\par

\section{Related Work}
\label{sec:related_work}

\subsection{Road and Infrastructure Benchmarks}
Vision-based inspection supports civil infrastructure monitoring and maintenance planning~\cite{Spencer_2019_InfrastructureInspection,Peraka_2020_PavementAsset,ElHakea_2023_PavementCV}, yet conventional methods and datasets predominantly target localized perception of cracks, potholes, guardrails, bridges, or roadway obstacles~\cite{Cha_2017_CrackDetection,Maeda_2018_RoadDamage,Fan_2020_PotholeSeg,Jin_2024_Guardrail,Zhang_2020_BridgeDamage,Zhou_2024_CrackSeg,Lis_2019_RoadAnomaly,Arya_2024_RDD2022,Zhu_2022_UAVPavement}. Beyond pure perception, general multimodal benchmarks evaluate broad visual reasoning~\cite{Liu_2024_MMBench,Yu_2024_MMVet,Yue_2024_MMMU,Fu_2025_VideoMME}, whereas industrial benchmarks progress from unsupervised defect detection~\cite{Bergmann_2019_MVTecAD,Zou_2022_VisA} to independent multi-task question answering in MMAD~\cite{Jiang_2025_MMAD}. In road scenarios, RoadBench introduces synthetic road-damage image and text pairs~\cite{Xiao_2026_RoadBench}, and DriveLM links perception, prediction, and planning through graph-structured QA~\cite{Sima_2024_DriveLM}. RoadSocial and RoadSceneBench address traffic-event analysis and relational road semantics~\cite{Parikh_2025_RoadSocial,Liu_2026_RoadSceneBench}. As summarized in Table~\ref{tab:comparison}, these benchmarks have limited coverage of highway infrastructure across Ground and UAV views and do not assess maintenance correctness under the diagnostic prerequisites of the HIIP-Chain. HighwayVerse bridges this gap with the HIIP-Chain spanning Anomaly Detection, Anomaly Recognition, Condition Analysis, and Maintenance Resolution.

\subsection{MLLMs for Anomaly Understanding}
Although general-purpose MLLMs couple visual encoders with language models for instruction following~\cite{Li_2023_BLIP2,Liu_2023_VisualInstructionTuning,Bai_2025_Qwen3VL,Wang_2025_InternVL35}, they still struggle with fine-grained visual discrimination~\cite{Tong_2024_MMVP,Xu_2025_AnomalyOV}. Existing remedies incorporate language-guided visual refinement~\cite{chen2024improving} or specialist localization modules~\cite{Gu_2024_AnomalyGPT,Li_2025_Myriad,Xu_2025_AnomalyOV}, which require task-specific training, while small-and-large model collaboration routes queries via lightweight modules to assist large models~\cite{chen2024data,chen2025improving}. KKA~\cite{chen2026kka} uses language model knowledge to generate anomalous examples for training category-level visual detectors. VISPROG and ViperGPT compose external visual tools without task-specific training~\cite{Gupta_2023_VisProg,Suris_2023_ViperGPT}, while MMAD retrieves textual domain knowledge~\cite{Jiang_2025_MMAD} and MuRAG learns multimodal retrieval and generation~\cite{Chen_2022_MuRAG}. For in-context example selection, Flamingo uses visual similarity~\cite{Alayrac_2022_Flamingo}, whereas MMICES adds text-based reranking~\cite{Chen_2025_MMICES}. Rather than flat similarity matching, \mbox{LaneRAG} constructs an offline hierarchical reference bank from coarse-grained to fine-grained anomaly categories and combines DINOv2 and CLIP representations~\cite{Oquab_2024_DINOv2,Radford_2021_CLIP} for route-guided retrieval across Primary, Expanded, and Global Scopes, assembling retrieved references into a reference-conditioned prompt without updating any MLLM parameters.

\section{HighwayVerse}
\label{sec:highwayverse}
\label{sec:benchmarks}

\subsection{HIIP-Chain Task Definition}
\label{sec:hv_tasks}

The HIIP-Chain organizes highway infrastructure inspection into a progressive framework connecting visual anomaly understanding with maintenance-oriented decisions across six questions in four stages in Figure~\ref{fig:highwayverse_overview}\subref{fig:highwayverse_overview_c}.

\begin{itemize}[leftmargin=*,topsep=2pt,itemsep=1pt,parsep=0pt]
\item \textbf{Stage 1 Anomaly Detection.}
In \textbf{T1 Anomaly Detection}, the model determines whether a highway infrastructure anomaly appears in the image.

\item \textbf{Stage 2 Anomaly Recognition.}
In \textbf{T2 Anomaly Recognition}, the model identifies the fine-grained anomaly category of the primary defect.

\item \textbf{Stage 3 Condition Analysis.}
Three parallel questions assess complementary defect attributes, where \textbf{T3 Anomaly Localization} determines spatial location relative to nearby components, \textbf{T4 Anomaly Description} characterizes visible appearance and features, and \textbf{T5 Risk Analysis} evaluates primary structural or traffic risk.

\item \textbf{Stage 4 Maintenance Resolution.}
In \textbf{T6 Maintenance Resolution}, the model selects an appropriate maintenance action to resolve the observed anomaly.
\end{itemize}

Across these four stages, answering subsequent stages builds upon correctly completing preceding stages, progressing from Anomaly Detection and Anomaly Recognition to Condition Analysis and Maintenance Resolution. Within Stage 3, T3--T5 operate in parallel because once Stage 2 establishes the anomaly category, spatial location, visible appearance and features, and safety risk capture three orthogonal dimensions of the same defect without mutual prerequisites, concurrently informing Stage 4 under the protocol in Section~\ref{sec:hv_protocol}.

\subsection{Dataset Construction}
\label{sec:hv_collection}

\begin{figure}[!t]
\centering
\includegraphics[
    width=\columnwidth,
    pagebox=mediabox,
    trim=162bp 55bp 162bp 55bp,
    clip
]{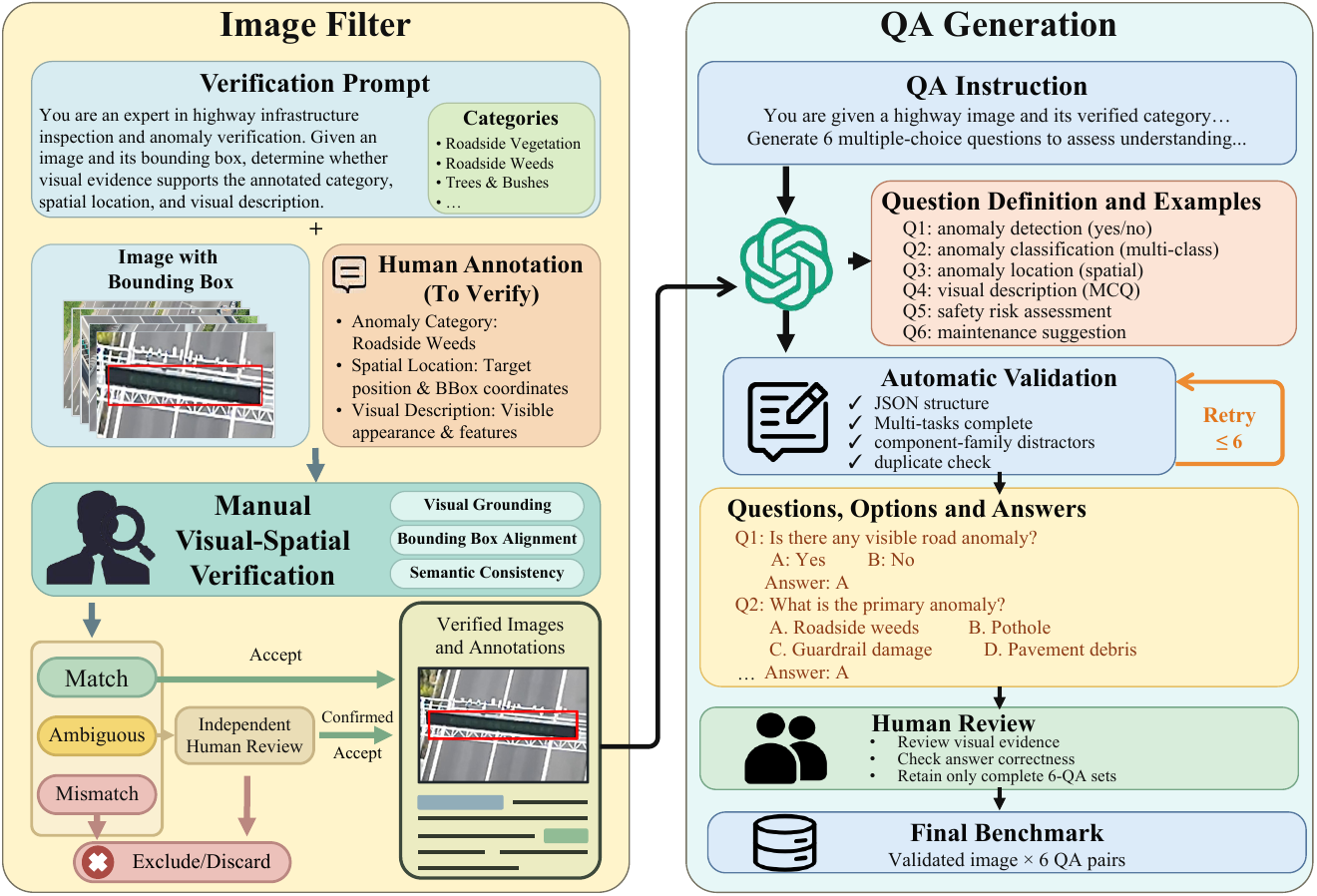}
\caption{HighwayVerse construction through image filtering, visual-spatial verification, QA generation, and human review.}
\label{fig:annotation_pipeline}
\end{figure}

\paragraph{Image Acquisition and Review.}
HighwayVerse is constructed from Ground and UAV imagery collected independently during six months of routine expressway and highway infrastructure inspection. Specifically, Ground captures target fine-grained defects on pavements, guardrails, glare screens, and drainage systems, whereas UAV views cover broader road corridors, slopes, and roadside surroundings. During collection, human inspectors also recorded field metadata specifying the anomaly category, spatial location, and visible appearance, while manually annotating target bounding boxes on every UAV record; Ground records require no bounding boxes because the target anomaly directly occupies the close-range subject. As illustrated in Figure~\ref{fig:annotation_pipeline}, an Image Filter first removed severely blurred, overly distant, duplicate, and near-duplicate records within and across splits, alongside UAV records lacking clean unboxed originals. Subsequently, Manual Visual-Spatial Verification checked Semantic Consistency against the human-annotated field metadata and Bounding Box Alignment, routing ambiguous cases to Independent Human Review while discarding mismatched, privacy-sensitive, or watermarked records.

\paragraph{Q\&A Annotation.}
\label{sec:hv_annotation}
Following Figure~\ref{fig:annotation_pipeline}, we prompted an MLLM, GPT-6.1-Terra, to generate six multiple-choice QA pairs per verified record from the human-annotated field metadata and HIIP-Chain task definitions. Questions Q1--Q6 in the prompt correspond to tasks T1--T6, respectively, where each QA pair contains a question, options, and a reference answer. T2 distractors were drawn from semantically related component-family categories, while T3--T6 distractors varied task-specific attributes. Automatic Validation combined with manual checks verified JSON structure, multi-task completeness, component-family distractors, duplicate removal, balanced answer positions, and consistent option lengths. In Human Review, 10 domain-trained annotators spent 400 hours reviewing visual evidence, answer correctness, and distractor plausibility, retaining only complete six-QA sets. The initial record-level pass rate was 82.3\%, and uncertain records underwent further manual review before retention. Appendix~\ref{app:complete_records} presents complete Ground and UAV examples with their original records, questions, options, and reference answers.

\paragraph{Dataset Statistics.}
\label{sec:hv_statistics}
Starting from 18,564 candidate records, image review, QA review, and split consolidation excluded 4,421 records, comprising 2,836 Ground and 1,585 UAV records, leaving 14,143 validated inspection records and 84,858 QA pairs. Table~\ref{tab:hv_dataset_statistics} summarizes the resulting dataset splits, which remain strictly image-disjoint within each viewpoint.

\begin{table}[!t]
\captionsetup{position=top,font=small}
\caption{HighwayVerse splits. Each record has six QA pairs.}
\label{tab:hv_dataset_statistics}
\centering
\small
\setlength{\tabcolsep}{3pt}
\begin{tabular}{lrrrrr}
\toprule
Viewpoint & Train & Val & Test & Records & QA Pairs \\
\midrule
Ground & 6,213 & 496 & 971 & 7,680 & 46,080 \\
UAV & 5,000 & 500 & 963 & 6,463 & 38,778 \\
\midrule
Total & 11,213 & 996 & 1,934 & 14,143 & 84,858 \\
\bottomrule
\end{tabular}
\end{table}

As shown in Figure~\ref{fig:highwayverse_overview}\subref{fig:category_distribution}, both viewpoints exhibit pronounced long-tailed category distributions, where the top three categories alone account for 24.38\% of all Ground records across 100 categories and 64.68\% of all UAV records across 17 categories, leaving a long spectrum of rare tail anomalies. During split construction, frequent categories were partitioned across training, validation, and test sets, whereas scarce tail categories were placed in the test sets to guarantee full-taxonomy evaluation. Consequently, the test sets cover all 100 Ground and 17 UAV categories, while the training sets contain 66 and 15 categories, respectively. Under this partitioning, the 34 test-only Ground categories account for 112 records, representing 11.5\% of the Ground test set, while the 2 test-only UAV categories contain one record each. Appendix~\ref{app:category_inventory} details the complete category inventory and per-split record counts.

\subsection{Chained Evaluation Protocol}
\label{sec:hv_protocol}
\label{sec:hv_metric}

To operationalize the HIIP-Chain as a rigorous evaluation standard, models answer all six questions per test record from the image and candidate options alone, without access to field metadata or ground-truth annotations. Conventional isolated scoring is vulnerable to superficial shortcuts, because an upstream question can be answered correctly by chance without genuine visual grounding. For instance, Stage 1 Anomaly Detection evaluates whether an anomaly exists, whereas Stage 2 Anomaly Recognition requires identifying the specific anomaly category, meaning a model may correctly guess the presence of a defect in Stage 1 while failing to identify the true target. By conditioning downstream progression along the HIIP-Chain on accurate upstream completions, our protocol filters out fortuitous upstream predictions and ensures that downstream success reflects authentic comprehension of highway infrastructure anomalies.

Specifically, let $c_{i,t}\in\{0,1\}$ denote whether the predicted option for question $t\in\{1,\dots,6\}$ on record $i$ matches the reference answer, with missing or invalid responses assigned zero. Along the HIIP-Chain, advancing to Stage 2 Anomaly Recognition requires a correct judgment in Stage 1 Anomaly Detection. Entering Stage 3 Condition Analysis requires correct predictions across both Stage 1 and Stage 2, while the three parallel condition tasks within Stage 3 are evaluated concurrently without mutual prerequisites. Culminating at Stage 4 Maintenance Resolution, credit is granted only when the model correctly completes Stage 1, Stage 2, at least one condition task in Stage 3, and Stage 4 itself. Formally, the chained score $s_{i,t}$ is defined as
\begin{equation}
s_{i,t}=
\begin{cases}
c_{i,1}, & t=1,\\
c_{i,1}c_{i,2}, & t=2,\\
c_{i,1}c_{i,2}c_{i,t}, & t\in\{3,4,5\},\\
c_{i,1}c_{i,2}c_{i,6}\,
\mathbf{1}\!\left[\sum_{j=3}^{5}c_{i,j}\geq1\right],
& t=6.
\end{cases}
\label{eq:dependency_gate}
\end{equation}

Over a test set $\mathcal{I}$, we quantify performance along the HIIP-Chain through per-question chained accuracy $\operatorname{Acc}_t$, the overall stage-spanning average $\operatorname{Mean}$, and our flagship end-to-end metric $\operatorname{Final}$, all expressed as percentages via
\begin{equation}
\begin{gathered}
\operatorname{Acc}_t
= \frac{100}{|\mathcal{I}|}
  \sum_{i\in\mathcal{I}}s_{i,t}, \\[4pt]
\operatorname{Mean}
= \frac{1}{6}\sum_{t=1}^{6}\operatorname{Acc}_t,
\quad
\operatorname{Final} = \operatorname{Acc}_6.
\end{gathered}
\label{eq:hv_metrics}
\end{equation}
While $\operatorname{Mean}$ summarizes retention across the HIIP-Chain, $\operatorname{Final}$ serves as our holistic end-to-end indicator, rewarding Maintenance Resolution only when backed by accurate upstream detection, recognition, and condition analysis.

\section{Method}
\label{sec:method}
\label{sec:method_task_scope}
\label{sec:method_ground}
\label{sec:method_routing}
\label{sec:method_agents}
\label{sec:method_inference}

\begin{figure}[t]
\centering
\includegraphics[width=\columnwidth,height=\columnwidth]{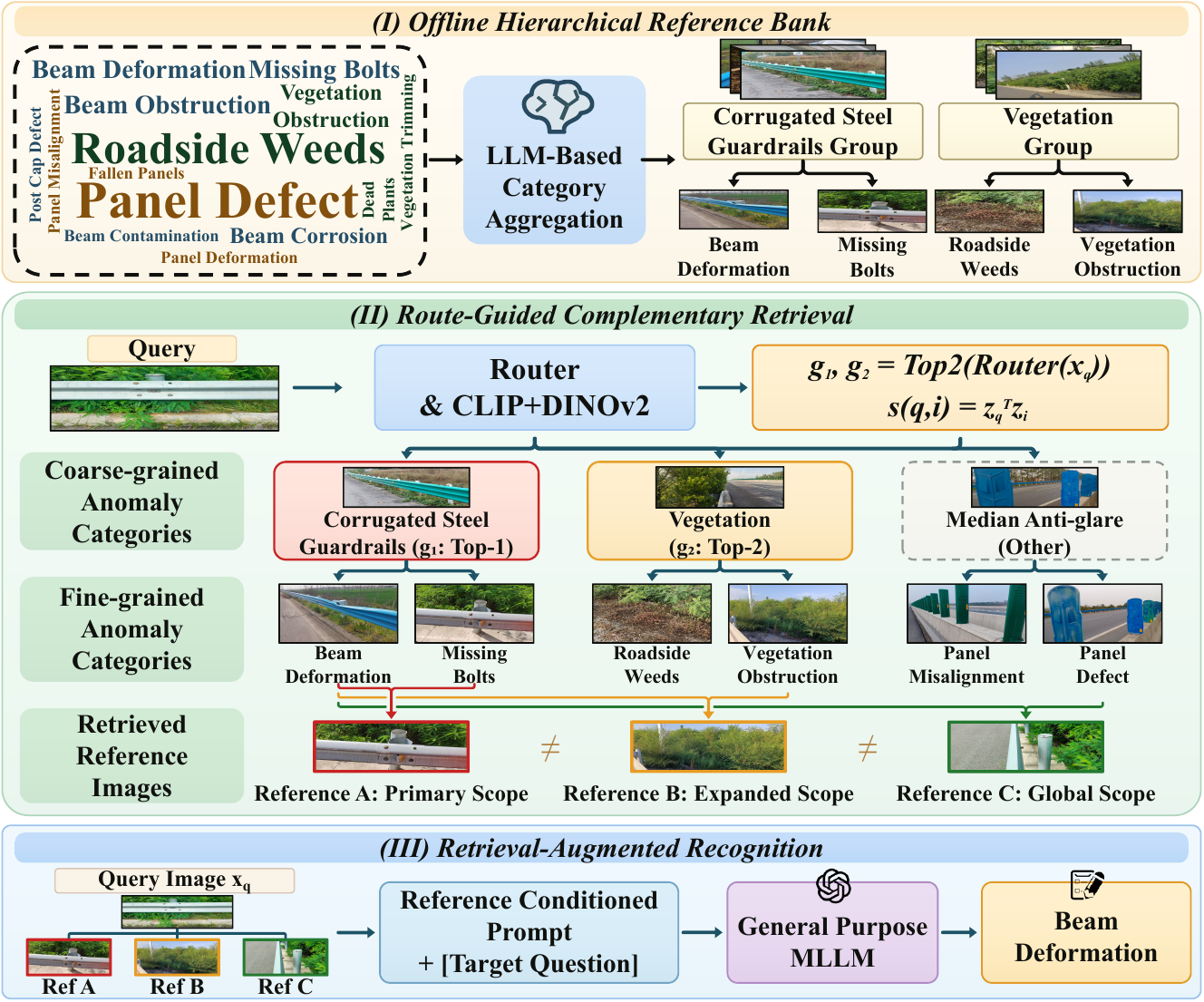}
\caption{\textbf{Overview of \mbox{LaneRAG}.} An offline hierarchical reference bank and router guide retrieval across Primary, Expanded, and Global Scopes to form a reference-conditioned prompt for frozen MLLMs.}
\label{fig:lanerag_pipeline}
\end{figure}

To strengthen highway infrastructure inspection along the HIIP-Chain without updating any MLLM parameters, we propose Local-and-Global Navigated Exemplar Retrieval-Augmented Guidance (\textbf{\mbox{LaneRAG}}), as illustrated in Figure~\ref{fig:lanerag_pipeline}. Operating as a companion training-free adaptation framework for frozen general-purpose MLLMs, \mbox{LaneRAG} organizes labeled training records into an offline hierarchical reference bank and retrieves complementary reference images across local and global scopes, supplying structured in-context visual guidance to ground anomaly discrimination and boost overall chained performance.

\subsection{Offline Hierarchical Reference Bank}
\label{sec:method_bank}

We construct viewpoint-specific offline hierarchical reference banks $\mathcal{B}$ for Ground and UAV imagery exclusively from the training splits in Table~\ref{tab:hv_dataset_statistics}. Rather than indexing flat anomaly labels directly, we apply LLM-based category aggregation to textual anomaly labels, consolidating diverse fine-grained anomaly categories into semantically coherent coarse-grained anomaly categories aligned with highway infrastructure entities. As shown in Figure~\ref{fig:lanerag_pipeline}, fine-grained categories such as Beam Deformation and Missing Bolts are aggregated under the Corrugated Steel Guardrails group, whereas Roadside Weeds and Vegetation Obstruction are grouped under the Vegetation group, yielding $G$ coarse-grained anomaly categories per viewpoint while preserving every record's original fine-grained label. Each training record is then indexed into $\mathcal{B}$ together with its visual embedding, original fine-grained anomaly category, and aggregated coarse-grained anomaly category, establishing a structured two-level reference hierarchy over $\mathcal{B}$ in which $\mathcal{B}_g \subset \mathcal{B}$ denotes the reference subset belonging to coarse-grained category $g$.

\subsection{Route-Guided Complementary Retrieval}
\label{sec:method_retrieval}

At inference, given a query image $x_q$, we extract complementary structural and semantic representations via frozen DINOv2 ViT-B/14~\cite{Oquab_2024_DINOv2} and CLIP ViT-B/32~\cite{Radford_2021_CLIP} encoders, denoted as $\phi_{\mathrm{DINO}}(x)\in\mathbb{R}^{d_{\mathrm{D}}}$ and $\phi_{\mathrm{CLIP}}(x)\in\mathbb{R}^{d_{\mathrm{C}}}$ with $d_{\mathrm{D}}=768$ and $d_{\mathrm{C}}=512$. Concatenating the independently $\ell_2$-normalized vectors forms a unified visual embedding $\mathbf{z}(x)\in\mathbb{R}^{1280}$ defined as
\begin{equation}
\mathbf{z}(x) = \frac{1}{\sqrt{2}}\left[\frac{\phi_{\mathrm{DINO}}(x)}{\|\phi_{\mathrm{DINO}}(x)\|_2} \,\Big\|\, \frac{\phi_{\mathrm{CLIP}}(x)}{\|\phi_{\mathrm{CLIP}}(x)\|_2}\right],
\label{eq:lanerag_embedding}
\end{equation}
which induces a balanced cosine similarity score $s(q,i)=\mathbf{z}(x_q)^\top\mathbf{z}_i$ between $x_q$ and any indexed reference $i\in\mathcal{B}$, and simultaneously precomputes the stored embeddings in $\mathcal{B}$.

To navigate the two-level reference hierarchy, a lightweight viewpoint-specific router, parameterized as an offline-trained two-layer MLP over standardized $\mathbf{z}(x_q)$, predicts a probability distribution across the $G$ coarse-grained anomaly categories and selects the top-two candidate groups $g_1$ and $g_2$. Guided by $g_1$, $g_2$, and $s(q,i)$, we retrieve three distinct reference images across progressively broader scopes via
\begin{equation}
\begin{aligned}
i_A &= \operatorname*{arg\,max}_{i \in \mathcal{B}_{g_1}}
s(q,i),\\
i_B &= \operatorname*{arg\,max}_{i \in
(\mathcal{B}_{g_1}\cup\mathcal{B}_{g_2})
\setminus\{i_A\}} s(q,i),\\
i_C &= \operatorname*{arg\,max}_{i \in
\mathcal{B}\setminus\{i_A,i_B\}}
s(q,i),
\end{aligned}
\label{eq:lanerag_routes}
\end{equation}
producing Reference A from the Primary Scope, Reference B from the Expanded Scope, and Reference C from the Global Scope.

\subsection{Retrieval-Augmented Recognition}
\label{sec:method_rag_classify}

During inference, the query image $x_q$ and recognition question are combined with Reference A, Reference B, and Reference C, their fine-grained anomaly category labels, and scope indicators into a reference-conditioned prompt for a frozen general-purpose MLLM. These progressive scopes play complementary discriminative roles, where Reference A from the Primary Scope typically anchors the dominant group with a closely matched candidate, Reference B from the Expanded Scope may introduce a confusable alternative from the runner-up group to sharpen boundary distinctions, and Reference C from the Global Scope often draws from broader or unrelated categories as a routing fallback or contrastive negative reference. Contrasting aligned, near-boundary, and distinct anomalies in context guides fine-grained discrimination without parameter updates. Integrating this guidance while holding remaining stages fixed improves overall chained performance under Equation~\ref{eq:dependency_gate}, with details in Appendix~\ref{app:rag_protocol}.

Appendix~\ref{app:full_taxonomy} evaluates the SAM3-based Segmentation-Guided Specialist Pipeline for full-taxonomy recognition across all 17 UAV anomaly categories.

\section{Experiments}
\label{sec:experiments}

\begin{table*}[t]
\centering
\captionsetup{position=top,font=small}
\caption{Baseline and LaneRAG results on HighwayVerse in \%. Gains are absolute improvements over each baseline. Final denotes chained Maintenance Resolution accuracy; Mean averages all six chained question accuracies.}
\label{tab:sequential_gated_results}
\label{tab:ground_results}
\label{tab:benchmark_results}
\fontsize{7.5}{9.3}\selectfont
\setlength{\tabcolsep}{1.6pt}
\renewcommand{\arraystretch}{1.18}
\definecolor{oursrow}{HTML}{E8F2FA}
\newcommand{\laneragcell}[2]{\shortstack[c]{#1\\[-1pt]{\fontsize{6}{6.5}\selectfont\color{black!65}(#2)}}}
\begin{tabular*}{\textwidth}{@{\extracolsep{\fill}}l*{7}{c}@{\hspace{5pt}}*{7}{c}@{\hspace{5pt}}>{\columncolor{blue!5}}c>{\columncolor{blue!5}}c@{}}
\toprule
& \multicolumn{7}{c}{\textbf{UAV ($N=963$)}} & \multicolumn{7}{c}{\textbf{Ground ($N=971$)}} & \multicolumn{2}{c}{\textbf{Overall}}\\
\cmidrule(lr){2-8}\cmidrule(lr){9-15}\cmidrule(l){16-17}
Model & Det. & Rec. & Loc. & Desc. & Risk & Final & Mean & Det. & Rec. & Loc. & Desc. & Risk & Final & Mean & Final & Mean\\
\midrule
\multicolumn{17}{@{}l}{\textit{Open-source models}} \\
Qwen2.5-VL-3B & 55.87 & 32.09 & 13.08 & 20.77 & 16.30 & 12.88 & 25.16 & 50.26 & 29.87 & 18.85 & 24.82 & 23.17 & 26.98 & 28.99 & 19.96 & 27.09 \\
\rowcolor{oursrow}
\quad + LaneRAG & \laneragcell{55.87}{+0.00} & \laneragcell{43.30}{+11.21} & \laneragcell{17.96}{+4.88} & \laneragcell{27.73}{+6.96} & \laneragcell{22.33}{+6.03} & \laneragcell{19.21}{+6.33} & \laneragcell{31.07}{+5.91} & \laneragcell{50.26}{+0.00} & \laneragcell{30.59}{+0.72} & \laneragcell{18.64}{-0.21} & \laneragcell{24.51}{-0.31} & \laneragcell{24.30}{+1.13} & \laneragcell{27.50}{+0.52} & \laneragcell{29.30}{+0.31} & \laneragcell{23.37}{+3.41} & \laneragcell{30.18}{+3.09} \\
Qwen2.5-VL-7B & 37.69 & 28.45 & 15.26 & 18.69 & 16.20 & 14.54 & 21.81 & 36.77 & 21.63 & 15.65 & 19.77 & 17.92 & 20.70 & 22.07 & 17.63 & 21.94 \\
\rowcolor{oursrow}
\quad + LaneRAG & \laneragcell{37.69}{+0.00} & \laneragcell{33.54}{+5.09} & \laneragcell{17.13}{+1.87} & \laneragcell{21.39}{+2.70} & \laneragcell{19.31}{+3.11} & \laneragcell{16.41}{+1.87} & \laneragcell{24.25}{+2.44} & \laneragcell{36.77}{+0.00} & \laneragcell{23.27}{+1.64} & \laneragcell{17.20}{+1.55} & \laneragcell{21.32}{+1.55} & \laneragcell{20.19}{+2.27} & \laneragcell{22.35}{+1.65} & \laneragcell{23.52}{+1.45} & \laneragcell{19.39}{+1.76} & \laneragcell{23.88}{+1.94} \\
InternVL3.5-8B & 72.48 & 52.34 & 24.09 & 25.34 & 28.66 & 25.96 & 38.14 & 71.78 & 45.31 & 28.63 & 31.31 & 31.31 & 40.89 & 41.54 & 33.45 & 39.85 \\
\rowcolor{oursrow}
\quad + LaneRAG & \laneragcell{72.48}{+0.00} & \laneragcell{64.69}{+12.35} & \laneragcell{30.11}{+6.02} & \laneragcell{30.84}{+5.50} & \laneragcell{34.58}{+5.92} & \laneragcell{32.71}{+6.75} & \laneragcell{44.24}{+6.10} & \laneragcell{71.78}{+0.00} & \laneragcell{47.27}{+1.96} & \laneragcell{30.59}{+1.96} & \laneragcell{31.82}{+0.51} & \laneragcell{32.03}{+0.72} & \laneragcell{42.53}{+1.64} & \laneragcell{42.67}{+1.13} & \laneragcell{37.64}{+4.19} & \laneragcell{43.45}{+3.60} \\
Qwen3-VL-8B & 54.00 & 44.55 & 25.96 & 33.23 & 30.32 & 29.08 & 36.19 & 56.13 & 39.75 & 31.31 & 36.97 & 37.49 & 38.72 & 40.06 & 33.92 & 38.13 \\
\rowcolor{oursrow}
\quad + LaneRAG & \laneragcell{54.00}{+0.00} & \laneragcell{51.40}{+6.85} & \laneragcell{28.76}{+2.80} & \laneragcell{37.49}{+4.26} & \laneragcell{34.79}{+4.47} & \laneragcell{32.40}{+3.32} & \laneragcell{39.81}{+3.62} & \laneragcell{56.13}{+0.00} & \laneragcell{40.37}{+0.62} & \laneragcell{31.82}{+0.51} & \laneragcell{37.59}{+0.62} & \laneragcell{38.11}{+0.62} & \laneragcell{39.55}{+0.83} & \laneragcell{40.59}{+0.53} & \laneragcell{35.99}{+2.07} & \laneragcell{40.20}{+2.07} \\
MiniCPM-V4.5 & 50.47 & 40.81 & 23.78 & 29.70 & 25.75 & 25.75 & 32.71 & 58.50 & 38.11 & 29.35 & 36.05 & 34.29 & 37.18 & 38.91 & 31.49 & 35.82 \\
\rowcolor{oursrow}
\quad + LaneRAG & \laneragcell{50.47}{+0.00} & \laneragcell{46.73}{+5.92} & \laneragcell{27.21}{+3.43} & \laneragcell{32.81}{+3.11} & \laneragcell{29.60}{+3.85} & \laneragcell{28.97}{+3.22} & \laneragcell{35.96}{+3.25} & \laneragcell{58.50}{+0.00} & \laneragcell{42.02}{+3.91} & \laneragcell{31.82}{+2.47} & \laneragcell{39.44}{+3.39} & \laneragcell{37.38}{+3.09} & \laneragcell{40.47}{+3.29} & \laneragcell{41.61}{+2.70} & \laneragcell{34.75}{+3.26} & \laneragcell{38.80}{+2.98} \\
Qwen3.5-9B & 54.00 & 44.86 & 30.74 & 37.18 & 36.14 & 31.78 & 39.11 & 32.13 & 22.25 & 17.92 & 21.52 & 21.42 & 22.04 & 22.88 & 26.89 & 30.96 \\
\rowcolor{oursrow}
\quad + LaneRAG & \laneragcell{54.00}{+0.00} & \laneragcell{50.78}{+5.92} & \laneragcell{34.58}{+3.84} & \laneragcell{41.74}{+4.56} & \laneragcell{39.98}{+3.84} & \laneragcell{35.10}{+3.32} & \laneragcell{42.70}{+3.59} & \laneragcell{32.13}{+0.00} & \laneragcell{22.45}{+0.20} & \laneragcell{18.33}{+0.41} & \laneragcell{21.73}{+0.21} & \laneragcell{21.63}{+0.21} & \laneragcell{22.25}{+0.21} & \laneragcell{23.09}{+0.21} & \laneragcell{28.65}{+1.76} & \laneragcell{32.85}{+1.89} \\
Qwen3.5-27B & 53.27 & 47.14 & 33.85 & 40.81 & 38.63 & 36.45 & 41.69 & 59.73 & 41.40 & 35.74 & 40.06 & 40.58 & 41.09 & 43.10 & 38.78 & 42.40 \\
\rowcolor{oursrow}
\quad + LaneRAG & \laneragcell{53.27}{+0.00} & \laneragcell{51.40}{+4.26} & \laneragcell{36.45}{+2.60} & \laneragcell{44.13}{+3.32} & \laneragcell{41.95}{+3.32} & \laneragcell{38.84}{+2.39} & \laneragcell{44.34}{+2.65} & \laneragcell{59.73}{+0.00} & \laneragcell{44.28}{+2.88} & \laneragcell{38.11}{+2.37} & \laneragcell{42.74}{+2.68} & \laneragcell{43.15}{+2.57} & \laneragcell{43.87}{+2.78} & \laneragcell{45.31}{+2.21} & \laneragcell{41.37}{+2.59} & \laneragcell{44.83}{+2.43} \\
Qwen3-VL-32B & 59.19 & 52.13 & 36.14 & 42.68 & 41.33 & 37.59 & 44.84 & 45.62 & 32.23 & 26.98 & 31.20 & 30.59 & 31.62 & 33.04 & 34.59 & 38.92 \\
\rowcolor{oursrow}
\quad + LaneRAG & \laneragcell{59.19}{+0.00} & \laneragcell{56.28}{+4.15} & \laneragcell{37.80}{+1.66} & \laneragcell{45.69}{+3.01} & \laneragcell{44.13}{+2.80} & \laneragcell{40.39}{+2.80} & \laneragcell{47.25}{+2.41} & \laneragcell{45.62}{+0.00} & \laneragcell{32.96}{+0.73} & \laneragcell{27.39}{+0.41} & \laneragcell{31.82}{+0.62} & \laneragcell{31.31}{+0.72} & \laneragcell{32.34}{+0.72} & \laneragcell{33.57}{+0.53} & \laneragcell{36.35}{+1.76} & \laneragcell{40.38}{+1.46} \\
\midrule
\multicolumn{17}{@{}l}{\textit{Closed-source models}} \\
GPT-4o & 65.42 & 56.80 & 30.11 & 42.37 & 46.42 & 39.15 & 46.71 & 73.94 & 51.08 & 40.78 & 48.61 & 48.20 & 50.77 & 52.23 & 44.98 & 49.48 \\
\rowcolor{oursrow}
\quad + LaneRAG & \laneragcell{65.42}{+0.00} & \laneragcell{58.88}{+2.08} & \laneragcell{31.26}{+1.15} & \laneragcell{43.72}{+1.35} & \laneragcell{48.08}{+1.66} & \laneragcell{40.39}{+1.24} & \laneragcell{47.96}{+1.25} & \laneragcell{73.94}{+0.00} & \laneragcell{53.55}{+2.47} & \laneragcell{42.33}{+1.55} & \laneragcell{50.77}{+2.16} & \laneragcell{50.57}{+2.37} & \laneragcell{53.35}{+2.58} & \laneragcell{54.09}{+1.86} & \laneragcell{46.90}{+1.92} & \laneragcell{51.03}{+1.55} \\
GPT-6.1-Terra & 67.19 & 62.51 & 55.04 & 58.36 & 59.40 & 57.84 & 60.06 & 75.39 & 59.42 & 54.17 & 58.29 & 58.50 & 59.11 & 60.81 & 58.48 & 60.44 \\
\rowcolor{oursrow}
\quad + LaneRAG & \laneragcell{67.19}{+0.00} & \laneragcell{64.07}{+1.56} & \laneragcell{56.39}{+1.35} & \laneragcell{60.02}{+1.66} & \laneragcell{60.64}{+1.24} & \laneragcell{58.77}{+0.93} & \laneragcell{61.18}{+1.12} & \laneragcell{75.39}{+0.00} & \laneragcell{59.73}{+0.31} & \laneragcell{54.58}{+0.41} & \laneragcell{58.50}{+0.21} & \laneragcell{58.60}{+0.10} & \laneragcell{59.42}{+0.31} & \laneragcell{61.04}{+0.23} & \laneragcell{59.10}{+0.62} & \laneragcell{61.11}{+0.67} \\
Gemini 2.5 Flash & 81.52 & 69.16 & 43.09 & 56.07 & 57.53 & 53.06 & 60.07 & 77.45 & 54.17 & 45.52 & 50.77 & 52.63 & 53.35 & 55.65 & 53.21 & 57.85 \\
\rowcolor{oursrow}
\quad + LaneRAG & \laneragcell{81.52}{+0.00} & \laneragcell{77.15}{+7.99} & \laneragcell{46.94}{+3.85} & \laneragcell{60.96}{+4.89} & \laneragcell{63.76}{+6.23} & \laneragcell{57.74}{+4.68} & \laneragcell{64.68}{+4.61} & \laneragcell{77.45}{+0.00} & \laneragcell{56.75}{+2.58} & \laneragcell{46.04}{+0.52} & \laneragcell{53.14}{+2.37} & \laneragcell{54.79}{+2.16} & \laneragcell{56.13}{+2.78} & \laneragcell{57.38}{+1.73} & \laneragcell{56.93}{+3.72} & \laneragcell{61.01}{+3.16} \\
Grok 4.20 Non-Reasoning & 68.22 & 55.56 & 29.80 & 39.25 & 38.84 & 33.13 & 44.13 & 59.84 & 41.09 & 28.94 & 38.93 & 37.59 & 39.55 & 40.99 & 36.35 & 42.55 \\
\rowcolor{oursrow}
\quad + LaneRAG & \laneragcell{68.22}{+0.00} & \laneragcell{57.53}{+1.97} & \laneragcell{30.32}{+0.52} & \laneragcell{39.98}{+0.73} & \laneragcell{40.19}{+1.35} & \laneragcell{33.54}{+0.41} & \laneragcell{44.96}{+0.83} & \laneragcell{59.84}{+0.00} & \laneragcell{42.43}{+1.34} & \laneragcell{30.07}{+1.13} & \laneragcell{40.06}{+1.13} & \laneragcell{38.93}{+1.34} & \laneragcell{40.89}{+1.34} & \laneragcell{42.04}{+1.05} & \laneragcell{37.23}{+0.88} & \laneragcell{43.49}{+0.94} \\
\bottomrule
\end{tabular*}
\end{table*}

\subsection{Experimental Setup}
\label{sec:exp_setup}

\paragraph{Evaluated Models.}
To assess multimodal capabilities across diverse architectures and parameter scales, we benchmark 12 representative MLLMs spanning 8 open-source models from 3B to 32B parameters, including Qwen2.5-VL~\cite{Bai_2025_Qwen25VL}, InternVL3.5~\cite{Wang_2025_InternVL35}, Qwen3-VL~\cite{Bai_2025_Qwen3VL}, MiniCPM-V4.5, and Qwen3.5, alongside 4 closed-source models, namely GPT-4o, GPT-6.1-Terra, Gemini 2.5 Flash, and Grok 4.20 Non-Reasoning. Every backbone is evaluated on the 971 Ground and 963 UAV test records both as a standalone baseline and with \mbox{LaneRAG}.

\paragraph{Implementation Details.}
We evaluate all models under the HIIP-Chain protocol in Section~\ref{sec:hv_protocol}, requesting single option-letter answers. Open-source models decode with temperature 0.7, top-$p$ 0.8, top-$k$ 20, and a 12-token output budget, while \mbox{LaneRAG} retrieves three reference images across the Primary, Expanded, and Global Scopes from the viewpoint-matched offline hierarchical reference banks of 6,213 Ground and 5,000 UAV training records. Predictions are scored by exact option matching over the full test-set size to report per-question chained accuracy, Mean, and Final, with Overall scores pooling both viewpoints by record count and further details in Appendix~\ref{app:eval_details}.

\subsection{Main Results}
\label{sec:exp_results}
\label{sec:exp_dependencies}
\label{sec:exp_framework}

Table~\ref{tab:benchmark_results} reports the performance of all 12 baseline MLLMs and their \mbox{LaneRAG}-adapted counterparts along the HIIP-Chain, yielding three findings.

\paragraph{Challenges of HighwayVerse.}
In Table~\ref{tab:benchmark_results}, all models attain relatively high accuracy in Stage 1 Anomaly Detection across both viewpoints yet drop steeply in Stage 2 Anomaly Recognition, showing that MLLMs fail to grasp fine-grained anomaly patterns and rely on false-positive Stage 1 guessing that inflates standalone detection reports. Consistently, Final falls well below Mean across every baseline. For example, Qwen2.5-VL-3B reaches 55.87\% on UAV and 50.26\% on Ground in Stage 1 Anomaly Detection with an Overall Mean of 27.09\%, yet drops to 32.09\% and 29.87\% in Stage 2 and only 19.96\% in Overall Final, while InternVL3.5-8B drops from over 71\% in Stage 1 to 33.45\% in Overall Final. These gaps confirm weak upstream anomaly comprehension and demonstrate the value of HIIP-Chain evaluation for infrastructure inspection.

\paragraph{Model Comparison.}
Comparing model families and scales shows how distinct capabilities govern sequential stages. In Stage 1 Anomaly Detection, differences remain modest across families and scales, with InternVL3.5-8B matching or exceeding GPT-6.1-Terra at 72.48\% versus 67.19\% on UAV and 71.78\% versus 75.39\% on Ground, and Qwen3.5-9B matching Qwen3.5-27B on UAV at 54.00\% versus 53.27\%. Clear gaps emerge in Stage 2 Anomaly Recognition, where GPT-6.1-Terra surpasses InternVL3.5-8B by 10.17\% on UAV and 14.11\% on Ground due to broader anomaly priors. Divergence widens further in Stage 3 Condition Analysis, where stronger logical and spatial reasoning lets GPT-6.1-Terra exceed InternVL3.5-8B by over 30\% in UAV Anomaly Localization at 55.04\% versus 24.09\% and Anomaly Description at 58.36\% versus 25.34\%, while Qwen3.5-27B similarly outpaces Qwen3.5-9B, yielding Overall Final scores of 58.48\% and 38.78\%. Distinct visual priors and analytical reasoning thus drive stage-wise divergence along the HIIP-Chain.

\paragraph{Effectiveness of LaneRAG.}
\mbox{LaneRAG} delivers universal gains across all 12 frozen MLLMs, raising Overall Mean by 2.15\% and Overall Final by 2.33\% on average, with substantially larger improvements on open-source than closed-source models. Specifically, open-source MLLMs gain averages of 2.43\% in Overall Mean, 2.60\% in Overall Final, and 6.97\% in UAV Stage 2 Anomaly Recognition, peaking at 12.35\% on InternVL3.5-8B, compared with 1.58\%, 1.78\%, and 3.40\% on closed-source models. Because open-source MLLMs carry fewer built-in priors over specialized highway anomalies, retrieved visual context more strongly compensates for missing domain knowledge and unlocks downstream HIIP-Chain credit.

\subsection{Ablation Study}
\label{sec:exp_rag_ablation}

\begin{table}[!t]
\centering
\captionsetup{position=top,font=small}
\caption{Ablation study of reference retrieval components on Stage 2 Anomaly Recognition chained accuracy in \%, with full \mbox{LaneRAG} highlighted in bold.}
\label{tab:ablation}
\label{tab:rag_ablation}
\footnotesize
\setlength{\tabcolsep}{2.2pt}
\renewcommand{\arraystretch}{1.1}

\begin{tabular*}{\columnwidth}{@{\extracolsep{\fill}}lccccc@{}}
\toprule
Model
& \shortstack{No\\RAG}
& \shortstack{w/o Global\\Scope}
& \shortstack{Out-of-Group\\Random}
& \shortstack{w/o Sim.\\Ranking}
& \shortstack{Lane\\RAG} \\
\midrule

\multicolumn{6}{@{}l}{\textit{UAV}\enspace($N=963$)} \\
Qwen2.5-VL-3B & 32.09 & 43.20 & 36.86 & 37.69 & \textbf{43.30} \\
Qwen2.5-VL-7B & 28.45 & 33.33 & 30.63 & 30.84 & \textbf{33.54} \\
InternVL3.5-8B & 52.34 & 63.45 & 58.36 & 59.50 & \textbf{64.69} \\
Qwen3-VL-8B & 44.55 & 50.99 & 47.87 & 48.49 & \textbf{51.40} \\
MiniCPM-V4.5 & 40.81 & 45.59 & 44.13 & 43.82 & \textbf{46.73} \\
Qwen3.5-9B & 44.86 & 50.78 & 47.14 & 47.25 & \textbf{50.99} \\
Qwen3.5-27B & 47.14 & 50.99 & 49.01 & 49.74 & \textbf{51.40} \\
Qwen3-VL-32B & 52.13 & 56.28 & 54.10 & 54.62 & \textbf{56.28} \\

\midrule
\multicolumn{6}{@{}l}{\textit{Ground}\enspace($N=971$)} \\
Qwen2.5-VL-3B & 29.87 & 29.76 & 28.63 & 28.84 & \textbf{29.87} \\
Qwen2.5-VL-7B & 21.63 & 22.04 & 20.91 & 20.70 & \textbf{23.27} \\
InternVL3.5-8B & 45.31 & 46.04 & 46.04 & 46.24 & \textbf{47.27} \\
Qwen3-VL-8B & 39.75 & 38.41 & 37.69 & 38.21 & \textbf{40.37} \\
MiniCPM-V4.5 & 38.11 & 39.55 & 38.11 & 38.00 & \textbf{42.02} \\
Qwen3.5-9B & 22.25 & 21.94 & 21.83 & 21.42 & \textbf{22.45} \\
Qwen3.5-27B & 41.40 & 42.33 & 42.43 & 43.25 & \textbf{44.28} \\
Qwen3-VL-32B & 32.23 & 32.03 & 32.03 & 31.62 & \textbf{32.96} \\

\bottomrule
\end{tabular*}
\end{table}

To assess each component under fixed Stage 1 pass lists, prompts, and seeds across eight open-source MLLMs, Table~\ref{tab:rag_ablation} compares \mbox{LaneRAG} with No RAG, \textit{w/o Global Scope} without Reference C, \textit{w/o Sim.\ Ranking} using random sampling within Primary, Expanded, and Global Scopes, and \textit{Out-of-Group Random} sampling Reference C outside the top-two groups. Full \mbox{LaneRAG} achieves the best or tied-best Stage 2 Anomaly Recognition accuracy across all 16 settings, gaining 6.99\% on UAV and 1.49\% on Ground over No RAG and confirming that every module contributes.

Module sensitivity differs markedly across viewpoints. Removing similarity ranking hurts UAV far more than Ground, dropping average accuracy by 3.30\% under \textit{w/o Sim.\ Ranking} and 3.78\% under \textit{Out-of-Group Random} on UAV versus 1.78\% and 1.85\% on Ground, with Qwen2.5-VL-3B falling 5.61\% from 43.30\% to 37.69\% and InternVL3.5-8B falling 5.19\% on UAV, while \textit{w/o Sim.\ Ranking} drops below No RAG on 6 of 8 Ground models, including 4.02\% on MiniCPM-V4.5, due to appearance-mismatched visual distraction. Conversely, \textit{w/o Global Scope} degrades the 100-category Ground set by 1.30\% on average, reaching 2.47\% from 42.02\% to 39.55\% on MiniCPM-V4.5 and falling below No RAG on four backbones, versus only 0.46\% on the UAV set, showing that Reference C from the Global Scope is vital for resolving cross-group ambiguity in large taxonomies.

\section{Conclusion}
\label{sec:conclusion}

In this work, we advance highway infrastructure inspection from isolated anomaly perception to the HIIP-Chain, a progressive chain structure linking Anomaly Detection and Anomaly Recognition with Condition Analysis and Maintenance Resolution. To support this paradigm, we construct HighwayVerse, a multi-view benchmark comprising 14,143 real-world inspection records and 84,858 expert-reviewed QA pairs across 7,680 Ground records in 100 anomaly categories and 6,463 UAV records in 17 anomaly categories under natural long-tailed distributions. Using HighwayVerse, we benchmark 12 open-source and closed-source MLLMs, showing that chained evaluation separates genuine anomaly comprehension from upstream guessing. Furthermore, we propose \mbox{LaneRAG}, a training-free adaptation framework tailored to HighwayVerse that consistently improves performance across diverse MLLMs without updating model parameters. Overall, HighwayVerse provides a rigorous foundation for developing trustworthy multimodal highway infrastructure inspection systems.

{\small
\bibliographystyle{ieeenat_fullname}
\bibliography{main}
}
\clearpage
\setcounter{table}{0}
\renewcommand{\thetable}{A\arabic{table}}
\renewcommand{\theHtable}{appendix.\arabic{table}}
\appendix
\makeatletter
\setlength{\@fptop}{0pt}
\setlength{\@dblfptop}{0pt}
\setlength{\@fpsep}{12pt}
\setlength{\@dblfpsep}{12pt}
\setlength{\@fpbot}{0pt plus 1fil}
\setlength{\@dblfpbot}{0pt plus 1fil}
\makeatother
\setcounter{figure}{0}
\renewcommand{\thefigure}{A\arabic{figure}}
\renewcommand{\theHfigure}{appendix.\arabic{figure}}

\section{HighwayVerse Data and Annotations}
\label{app:data_sources}

\paragraph{Acquisition and Viewpoint Characteristics.}
We collected Ground photographs and UAV images from operational expressway and arterial highway infrastructure inspection programs over a six-month period. Ground photographs, acquired manually near component height, capture surface damage, structural connections, and fine-grained local defects. UAV imagery covers wider road segments, slopes, and roadside corridors, placing anomalies in spatial context for risk assessment and Maintenance Resolution while posing challenges from overhead views, shadows, and small targets.

\paragraph{Image Curation and Review.}
Image filtering, verification, and split consolidation reduced 18,564 candidate records to 14,143 verified records comprising 7,680 Ground and 6,463 UAV records. Exact-pixel and visual-similarity checks removed duplicate observations, while Manual Visual-Spatial Verification and screening excluded unsuitable captures, annotation conflicts, and identifying overlays, as detailed in Section~\ref{sec:hv_collection}. A team of 10 domain-trained annotators spent 400 hours reviewing the QA annotations, achieving an initial record-level pass rate of 82.3\% with further manual review on uncertain records.

\subsection{Complete Inspection Records}
\label{app:complete_records}
\begin{figure*}[!t]
\centering
\newcommand{\hvrecordimage}[1]{%
  \begingroup
  \setlength{\fboxsep}{0pt}%
  \setlength{\fboxrule}{0.3pt}%
  \fcolorbox{black!20}{white}{%
    \parbox[c][40mm][c]{52mm}{%
      \centering\includegraphics[width=\linewidth,keepaspectratio]{#1}}}%
  \endgroup}
\begin{minipage}[t]{0.485\textwidth}
\vspace{0pt}
\centering\textbf{(a) Ground}\par
{\scriptsize road\_030056 $\mid$ Pavement Pothole}\par
\vspace{3pt}
\hvrecordimage{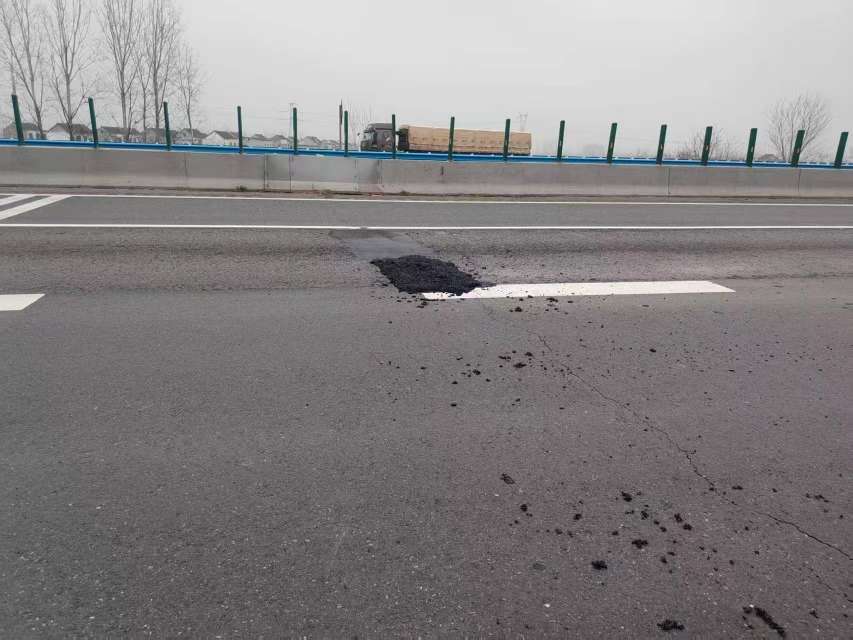}
\par\vspace{4pt}\raggedright\fontsize{8}{9.2}\selectfont
\setlength{\parindent}{0pt}\setlength{\parskip}{3pt}
\textbf{T1: Anomaly Detection.} Is there any visible highway anomaly in the image?\par\nopagebreak
\textbf{A.} Yes\quad\textbf{B.} No\par
\textit{Answer: A.}\par
\textbf{T2: Anomaly Recognition.} What type of pavement anomaly is visible where the white lane marking is interrupted in the middle foreground?\par\nopagebreak
{\setlength{\parskip}{0pt}
\textbf{A.} Missing or Defective Pavement Marking\par
\textbf{B.} Pavement Debris\par
\textbf{C.} Longitudinal Pavement Crack\par
\textbf{D.} Pavement Pothole\par
}
\textit{Answer: D.}\par
\textbf{T3: Anomaly Localization.} Where is the primary pavement defect positioned relative to the visible lane markings?\par\nopagebreak
{\setlength{\parskip}{0pt}
\textbf{A.} On the pavement at the middle-foreground white lane marking\par
\textbf{B.} On the pavement beside the left foreground marking segment\par
\textbf{C.} On the pavement along the upper lane line near the barrier\par
\textbf{D.} On the pavement in the lower-right foreground wheel path\par
}
\textit{Answer: A.}\par
\textbf{T4: Anomaly Description.} Which visible pavement condition best characterizes the damaged area at the white marking?\par\nopagebreak
{\setlength{\parskip}{0pt}
\textbf{A.} The pavement has a narrow crack running diagonally toward the lower right\par
\textbf{B.} The pavement has a broad dark surface stain extending across the lane\par
\textbf{C.} The pavement has a localized irregular cavity with broken dark asphalt edges\par
\textbf{D.} The pavement has scattered loose fragments without a damaged surface area\par
}
\textit{Answer: C.}\par
\textbf{T5: Risk Analysis.} What is the most directly supported traffic risk from the localized damaged pavement at the lane marking?\par\nopagebreak
{\setlength{\parskip}{0pt}
\textbf{A.} The pavement crack may gradually widen under repeated traffic loading\par
\textbf{B.} The pavement cavity may cause a tire impact or steering disturbance when crossed\par
\textbf{C.} The pavement marking may become harder to distinguish in low light\par
\textbf{D.} The pavement surface may lose traction if a thin contaminant film develops\par
}
\textit{Answer: B.}\par
\textbf{T6: Maintenance Resolution.} Which pavement treatment best fits the localized broken area crossing the white lane marking?\par\nopagebreak
{\setlength{\parskip}{0pt}
\textbf{A.} Seal the narrow pavement cracks while leaving the broken area in place\par
\textbf{B.} Mill and resurface the full visible pavement lane length\par
\textbf{C.} Remove loose pavement material, prepare the edges, and install a localized asphalt patch\par
\textbf{D.} Clean the pavement and repaint only the interrupted white marking\par
}
\textit{Answer: C.}\par
\end{minipage}
\hfill
\begin{minipage}[t]{0.485\textwidth}
\vspace{0pt}
\centering\textbf{(b) UAV}\par
{\scriptsize JTA202609030903094917 $\mid$ Missing Side Ditch Cover}\par
\vspace{3pt}
\hvrecordimage{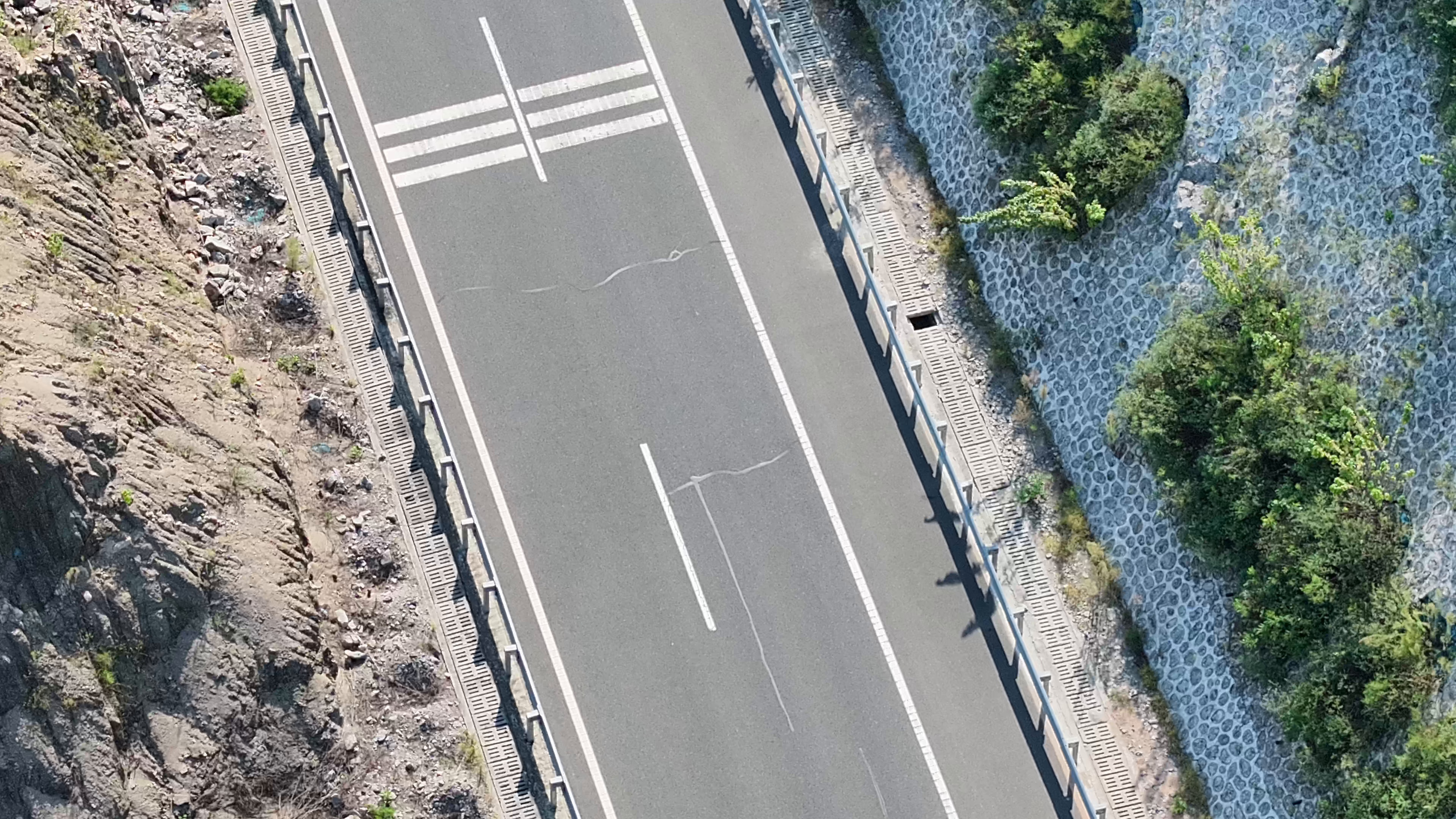}
\par\vspace{4pt}\raggedright\fontsize{8}{9.2}\selectfont
\setlength{\parindent}{0pt}\setlength{\parskip}{3pt}
\textbf{T1: Anomaly Detection.} Is there any visible highway anomaly in the image?\par\nopagebreak
\textbf{A.} Yes\quad\textbf{B.} No\par
\textit{Answer: A.}\par
\textbf{T2: Anomaly Recognition.} What type of anomaly is visible in the right-side drainage-cover run beside the guardrail?\par\nopagebreak
{\setlength{\parskip}{0pt}
\textbf{A.} Broken Side Ditch Cover\par
\textbf{B.} Debris on Side Ditch Cover\par
\textbf{C.} Missing Side Ditch Cover\par
\textbf{D.} Displaced Side Ditch Cover\par
}
\textit{Answer: C.}\par
\textbf{T3: Anomaly Localization.} Where is the interrupted drainage-cover section located in the full aerial view?\par\nopagebreak
{\setlength{\parskip}{0pt}
\textbf{A.} On the left-side ditch beside the upper-middle guardrail\par
\textbf{B.} On the left-side ditch beside the lower-middle guardrail\par
\textbf{C.} On the right-side ditch beside the lower-middle guardrail\par
\textbf{D.} On the right-side ditch beside the upper-middle guardrail\par
}
\textit{Answer: D.}\par
\textbf{T4: Anomaly Description.} Which detail best characterizes the visible condition within the right-side ditch-cover line?\par\nopagebreak
{\setlength{\parskip}{0pt}
\textbf{A.} The ditch-cover line has one slotted panel obscured by dark material\par
\textbf{B.} The ditch-cover line has one raised slab resting above the slotted run\par
\textbf{C.} The ditch-cover line has one fractured slab retaining light-colored pieces\par
\textbf{D.} The ditch-cover line has one dark rectangular interruption aligned with the slotted run\par
}
\textit{Answer: D.}\par
\textbf{T5: Risk Analysis.} Which conditional drainage consequence is most directly associated with the visible right-side ditch-cover condition?\par\nopagebreak
{\setlength{\parskip}{0pt}
\textbf{A.} The broken side-ditch cover could release fragments into the channel\par
\textbf{B.} The displaced side-ditch cover could form a raised edge beside the guardrail\par
\textbf{C.} The exposed side-ditch opening could admit debris that restricts local flow\par
\textbf{D.} The debris-covered side-ditch cover could reduce inflow through its slots\par
}
\textit{Answer: C.}\par
\textbf{T6: Maintenance Resolution.} Which replacement extent best fits the localized right-side ditch-cover condition beside the guardrail?\par\nopagebreak
{\setlength{\parskip}{0pt}
\textbf{A.} Inspect bearing edges and replace the two adjacent ditch covers\par
\textbf{B.} Inspect bearing edges and replace the three-cover ditch span\par
\textbf{C.} Inspect bearing edges and replace the offset neighboring ditch cover\par
\textbf{D.} Inspect bearing edges and replace the single affected ditch cover\par
}
\textit{Answer: D.}\par
\end{minipage}
\caption{Two complete HighwayVerse records. Original question wording, options, and reference answer keys are retained; task headings use the manuscript terminology. The Ground example is from the training split and the UAV example from the test split.}
\label{fig:hv_complete_records}
\end{figure*}

Figure~\ref{fig:hv_complete_records} presents a Ground record and a UAV record with their original images and complete six-question annotations. T1 has two options; T2--T6 have four. The T2 alternatives remain within the pavement family for the Ground example and the ditch-cover family for the UAV example.

\subsection{Complete Category Inventories}
\label{app:category_inventory}
Tables~\ref{tab:hv_ground_categories} and~\ref{tab:hv_uav_categories} enumerate the 100 Ground and 17 UAV categories under their respective source taxonomies. These are viewpoint-specific inventories, rather than a merged set of 117 mutually exclusive semantic categories. Counts cover the current train, validation, and test splits, where a dagger marks categories with fewer than 10 records across all three splits. Ground training and validation cover 66 and 53 categories, respectively, while UAV training and validation cover 15 and 12, and both test sets cover their complete source inventories.

\begin{table*}[!t]
\centering
\caption{All 100 Ground categories, in alphabetical order. Tr./V./Te. denote train/validation/test counts; All is their sum. $^{\dagger}$Fewer than 10 records across the full Ground subset. The two panels list all 7,680 records: 6,213 train, 496 validation, and 971 test.}
\label{tab:hv_ground_categories}
\begingroup
\fontsize{8}{9.2}\selectfont
\setlength{\tabcolsep}{2.4pt}
\renewcommand{\arraystretch}{1.05}
\begin{minipage}[t]{0.49\textwidth}
\vspace{0pt}
\begin{tabular}{@{}p{0.68\linewidth}rrrr@{}}
\toprule
Category & Tr. & V. & Te. & All \\
\midrule
Blocked Drainage Facility & 263 & 21 & 16 & 300 \\
Climbing Vegetation on Barbed Wire Fence$^{\dagger}$ & 0 & 0 & 1 & 1 \\
Climbing Vegetation on Isolation Fence & 44 & 6 & 13 & 63 \\
Climbing Vegetation on Noise Barrier$^{\dagger}$ & 0 & 0 & 4 & 4 \\
Collapsed Anti-glare Panel & 58 & 4 & 11 & 73 \\
Collapsed Barbed Wire Fence & 44 & 2 & 13 & 59 \\
Collapsed Crash Barrel$^{\dagger}$ & 0 & 0 & 1 & 1 \\
Collapsed Isolation Fence & 105 & 9 & 13 & 127 \\
Collapsed Water-filled Barrier$^{\dagger}$ & 0 & 0 & 2 & 2 \\
Contaminated Anti-glare Panel$^{\dagger}$ & 0 & 0 & 1 & 1 \\
Contaminated Guardrail Beam & 5 & 3 & 12 & 20 \\
Contaminated Noise Barrier & 17 & 3 & 12 & 32 \\
Corroded Guardrail Beam & 58 & 4 & 13 & 75 \\
Corroded Guardrail Beam Bolt$^{\dagger}$ & 0 & 0 & 1 & 1 \\
Corroded Guardrail Post$^{\dagger}$ & 0 & 0 & 7 & 7 \\
Corroded Isolation Fence Post$^{\dagger}$ & 0 & 0 & 4 & 4 \\
Corroded Noise Barrier$^{\dagger}$ & 0 & 0 & 2 & 2 \\
Culvert$^{\dagger}$ & 0 & 0 & 2 & 2 \\
Damaged Anti-glare Panel & 3 & 0 & 12 & 15 \\
Damaged Anti-throw Screen & 1 & 0 & 12 & 13 \\
Damaged Barbed Wire Fence & 88 & 7 & 13 & 108 \\
Damaged Bridge Structure & 3 & 0 & 12 & 15 \\
Damaged Concrete Barrier & 85 & 11 & 13 & 109 \\
Damaged Crash Barrel$^{\dagger}$ & 0 & 0 & 6 & 6 \\
Damaged Crash Cushion$^{\dagger}$ & 1 & 0 & 5 & 6 \\
Damaged Culvert Wing Wall$^{\dagger}$ & 0 & 0 & 1 & 1 \\
Damaged Curb & 149 & 12 & 14 & 175 \\
Damaged Drainage Facility & 406 & 35 & 21 & 462 \\
Damaged Expansion Joint & 312 & 18 & 19 & 349 \\
Damaged Guardrail Post$^{\dagger}$ & 0 & 0 & 9 & 9 \\
Damaged Isolation Fence & 43 & 5 & 12 & 60 \\
Damaged Isolation Fence Post$^{\dagger}$ & 1 & 0 & 8 & 9 \\
Damaged Noise Barrier$^{\dagger}$ & 0 & 0 & 5 & 5 \\
Damaged Passage Canopy$^{\dagger}$ & 0 & 0 & 4 & 4 \\
Damaged Pavement & 43 & 3 & 13 & 59 \\
Damaged Retaining Wall & 10 & 3 & 13 & 26 \\
Damaged Road Shoulder & 245 & 13 & 15 & 273 \\
Damaged Slope Protection & 98 & 11 & 13 & 122 \\
Damaged Traffic Sign$^{\dagger}$ & 0 & 0 & 4 & 4 \\
Damaged Traffic Sign Post$^{\dagger}$ & 0 & 0 & 1 & 1 \\
Dead Plant & 62 & 3 & 13 & 78 \\
Debris in Culvert or Passage$^{\dagger}$ & 1 & 1 & 7 & 9 \\
Debris in Drainage Facility & 98 & 9 & 12 & 119 \\
Debris in Expansion Joint & 92 & 8 & 13 & 113 \\
Debris on Bridge Structure & 0 & 0 & 10 & 10 \\
Deformed Anti-glare Panel & 6 & 1 & 12 & 19 \\
Deformed Bridge Bearing$^{\dagger}$ & 0 & 0 & 1 & 1 \\
Deformed Guardrail Beam & 151 & 16 & 13 & 180 \\
Deformed Noise Barrier & 54 & 3 & 13 & 70 \\
Exposed Reinforcement in Bridge Structure & 1 & 0 & 10 & 11 \\
\bottomrule\end{tabular}\end{minipage}\hfill%
\begin{minipage}[t]{0.49\textwidth}
\vspace{0pt}
\begin{tabular}{@{}p{0.68\linewidth}rrrr@{}}
\toprule
Category & Tr. & V. & Te. & All \\
\midrule
Fallen Tree & 2 & 0 & 11 & 13 \\
Fallen Vegetation$^{\dagger}$ & 0 & 0 & 1 & 1 \\
Improperly Positioned Crash Barrel$^{\dagger}$ & 1 & 0 & 7 & 8 \\
Longitudinal Pavement Crack & 102 & 12 & 12 & 126 \\
Misaligned Anti-glare Panel & 120 & 6 & 13 & 139 \\
Missing Guardrail Beam Bolt & 95 & 10 & 13 & 118 \\
Missing or Defective Anti-glare Panel & 410 & 34 & 21 & 465 \\
Missing or Defective Anti-throw Screen$^{\dagger}$ & 0 & 0 & 4 & 4 \\
Missing or Defective Barbed Wire Fence$^{\dagger}$ & 0 & 0 & 8 & 8 \\
Missing or Defective Delineator & 111 & 7 & 13 & 131 \\
Missing or Defective Guardrail Beam$^{\dagger}$ & 0 & 0 & 5 & 5 \\
Missing or Defective Guardrail Blockout & 1 & 0 & 9 & 10 \\
Missing or Defective Guardrail Post & 2 & 0 & 11 & 13 \\
Missing or Defective Guardrail Post Cap & 33 & 3 & 13 & 49 \\
Missing or Defective Isolation Fence & 55 & 8 & 13 & 76 \\
Missing or Defective Isolation Fence Post & 5 & 2 & 12 & 19 \\
Missing or Defective Noise Barrier & 51 & 3 & 13 & 67 \\
Missing or Defective Pavement Marking & 64 & 3 & 12 & 79 \\
Missing or Defective Traffic Sign Post$^{\dagger}$ & 0 & 0 & 5 & 5 \\
Missing or Defective Tunnel Wall Tile$^{\dagger}$ & 0 & 0 & 5 & 5 \\
Missing Traffic Sign$^{\dagger}$ & 0 & 0 & 1 & 1 \\
Missing Tree & 4 & 0 & 12 & 16 \\
Missing Yellow Flashing Light$^{\dagger}$ & 0 & 0 & 1 & 1 \\
Obstructed Barbed Wire Fence & 64 & 4 & 13 & 81 \\
Obstructed Guardrail Beam & 93 & 7 & 13 & 113 \\
Obstructed Isolation Fence & 212 & 16 & 15 & 243 \\
Obstructed Noise Barrier$^{\dagger}$ & 0 & 0 & 1 & 1 \\
Obstructed Traffic Sign & 6 & 0 & 11 & 17 \\
Other$^{\dagger}$ & 0 & 0 & 4 & 4 \\
Pavement Contamination & 6 & 2 & 12 & 20 \\
Pavement Debris & 141 & 6 & 15 & 162 \\
Pavement Pothole & 735 & 55 & 29 & 819 \\
Road Shoulder Debris & 95 & 6 & 13 & 114 \\
Roadside Weeds & 521 & 42 & 25 & 588 \\
Rockfall on Slope$^{\dagger}$ & 0 & 0 & 1 & 1 \\
Slope Collapse & 24 & 3 & 13 & 40 \\
Slope Debris & 202 & 9 & 16 & 227 \\
Snow-covered Pavement & 1 & 1 & 10 & 12 \\
Standing Water on Pavement & 11 & 4 & 12 & 27 \\
Tilted Kilometer Marker$^{\dagger}$ & 0 & 0 & 3 & 3 \\
Tilted Traffic Sign$^{\dagger}$ & 1 & 0 & 8 & 9 \\
Transverse Pavement Crack & 12 & 3 & 10 & 25 \\
Unclean Pavement & 296 & 19 & 19 & 334 \\
Undermined Slope$^{\dagger}$ & 0 & 0 & 2 & 2 \\
Vegetation Obstruction & 101 & 7 & 12 & 120 \\
Vegetation Requiring Trimming & 56 & 7 & 11 & 74 \\
Water Seepage on Tunnel Wall$^{\dagger}$ & 0 & 0 & 2 & 2 \\
Water-damaged Cone Slope & 13 & 3 & 13 & 29 \\
Water-damaged Road Shoulder & 20 & 3 & 13 & 36 \\
Water-damaged Slope$^{\dagger}$ & 0 & 0 & 3 & 3 \\
\bottomrule\end{tabular}\end{minipage}

\endgroup
\end{table*}

\begin{table}[!t]
\centering
\caption{All 17 UAV categories. $^{\dagger}$Fewer than 10 UAV records.}
\label{tab:hv_uav_categories}
\fontsize{8}{9.2}\selectfont
\setlength{\tabcolsep}{2.4pt}
\begin{tabular*}{\columnwidth}{@{\extracolsep{\fill}}p{0.58\columnwidth}rrrr@{}}
\toprule
Category & Train & Val & Test & All \\
\midrule
Anti-glare Panel Anomaly & 8 & 10 & 12 & 30 \\
Damaged Guardrail$^{\dagger}$ & 1 & 0 & 1 & 2 \\
Damaged Traffic Sign & 280 & 26 & 52 & 358 \\
Falling Object Protection Fence Anomaly$^{\dagger}$ & 1 & 0 & 2 & 3 \\
Fire$^{\dagger}$ & 0 & 0 & 1 & 1 \\
Missing Side Ditch Cover & 1277 & 121 & 217 & 1615 \\
Parked Construction Vehicle & 596 & 69 & 101 & 766 \\
Parked Vehicle & 201 & 26 & 37 & 264 \\
Pedestrian & 126 & 9 & 28 & 163 \\
Road Debris & 1174 & 82 & 178 & 1434 \\
Road Worker & 883 & 83 & 165 & 1131 \\
Slope Drainage System Anomaly & 57 & 10 & 25 & 92 \\
Smoke$^{\dagger}$ & 0 & 0 & 1 & 1 \\
Traffic Collision & 1 & 0 & 13 & 14 \\
Traffic Congestion & 19 & 9 & 12 & 40 \\
Variable Message Sign Anomaly & 189 & 16 & 33 & 238 \\
Water Erosion of Slope & 187 & 39 & 85 & 311 \\
\midrule
Total & 5,000 & 500 & 963 & 6,463 \\
\bottomrule
\end{tabular*}
\end{table}

\section{LaneRAG Implementation}
\label{app:rag_protocol}
\paragraph{Reference Banks.}
Separate offline hierarchical reference banks contain 6,213 Ground training records and 5,000 UAV training records. As described in Section~\ref{sec:method_bank}, LLM-based category aggregation consolidates fine-grained textual anomaly labels into 15 Ground and four UAV coarse-grained anomaly categories aligned with highway infrastructure entities. Each indexed record stores its image, aggregated coarse-grained anomaly category, original fine-grained anomaly category, and fused visual embedding. The coarse-grained anomaly category supervises router training, whereas the fine-grained anomaly category supplies the label shown in context to the MLLM. Retrieval uses the normalized fusion embedding in Equation~\ref{eq:lanerag_embedding}, whereas the router takes its standardized counterpart.

\paragraph{Router Implementation Details.}
\label{app:router_implementation}
The Ground router uses 6,213 training and 496 validation records, while the UAV router uses 5,000 training and 500 validation records. Feature-standardization statistics and inverse-frequency class weights are estimated from training records only. Each MLP is trained with AdamW, a learning rate of $10^{-3}$, weight decay of $10^{-4}$, batch size 64, and dropout 0.2. Training uses seed 42 and runs for at most 100 epochs, with early stopping after 15 epochs without improvement. Checkpoints are selected by validation macro-F1 over coarse-grained anomaly categories, with ties broken by lower validation loss. The selected Ground and UAV checkpoints correspond to epochs 19 and 7, respectively. During retrieval-augmented inference, router and encoder parameters remain frozen. Training-free MLLM adaptation means that the downstream MLLM receives retrieved references without parameter updates, while the auxiliary routers are trained offline.

\paragraph{Reference Retrieval.}
The viewpoint-specific router selects the query's top-two coarse-grained anomaly categories, and cosine similarity ranks the training records. Reference A in the Primary Scope is the nearest record within the top-1 group. Reference B in the Expanded Scope is the nearest remaining record within the union of the top-two groups. Reference C in the Global Scope is the nearest remaining record across the full reference bank. This procedure returns three distinct reference images with their fine-grained anomaly category labels and scope indicators in a reference-conditioned prompt, followed by the query image and Stage 2 Anomaly Recognition question with candidate options, returning a single option letter without parameter updates.

\paragraph{Reference-Selection Ablations.}
Table~\ref{tab:rag_ablation} compares full \mbox{LaneRAG} against \textit{No RAG} and three ablated variants. \textit{w/o Global Scope} removes Reference C from the Global Scope while retaining Reference A and Reference B from the Primary and Expanded Scopes. \textit{w/o Sim.\ Ranking} randomly samples three distinct labeled references from the exact same Primary, Expanded, and Global Scopes without cosine similarity ranking. \textit{Out-of-Group Random} keeps random sampling for the first two scopes but draws the third reference strictly outside the top-two routed groups. All variants share preprocessing, prompts, Stage 1 pass lists, and per-record generation seeds.

\paragraph{Stage 2 Replacement and Chained Evaluation.}
For each test record, \mbox{LaneRAG} Stage 2 credit requires both the original Stage 1 answer and the new Stage 2 answer to be correct. Original answers for Stage 1, Stage 3, and Stage 4 are held fixed, and chained scores are recomputed under Equation~\ref{eq:dependency_gate} in Section~\ref{sec:hv_protocol}.

\section{Evaluation Setup and Response Handling}
\label{app:eval_details}

\paragraph{Evaluation Sets and Response Scoring.}
Evaluation uses the fixed test sets of 971 Ground and 963 UAV records specified in Section~\ref{sec:hv_statistics}, with fixed membership across models. Inputs comprise task questions, candidate options, and the images permitted by each setting. Open-source decoding uses temperature 0.7, top-$p$ 0.8, top-$k$ 20, and a 12-token output limit, following Section~\ref{sec:exp_setup}. Selected options are scored by exact matching, where missing or invalid answers, refusals, and coordinate-only responses count as errors. Ordinary prediction errors do not trigger retries.

\paragraph{Evaluated Models.}
The main HIIP-Chain comparison includes 12 MLLMs spanning 8 open-source models and 4 closed-source models, namely GPT-4o, GPT-6.1-Terra, Gemini 2.5 Flash, and Grok 4.20 Non-Reasoning. Each backbone is evaluated both as a standalone baseline and with \mbox{LaneRAG} across both viewpoints in Table~\ref{tab:benchmark_results}. The full-taxonomy model roster is specified in Appendix~\ref{app:full_taxonomy}.

\section{Additional Benchmark Analyses}
\label{app:benchmark_analyses}

\paragraph{Evaluation Scope.}
\label{app:diagnostic_records}
The analyses below use the 971 Ground and 963 UAV test records evaluated in Table~\ref{tab:benchmark_results}, covering all 100 Ground and 17 UAV anomaly categories across the same 8 open-source and 4 closed-source models under the HIIP-Chain protocol in Section~\ref{sec:hv_protocol}.

\paragraph{Stage Prerequisites and Condition Analysis.}
\label{app:error_stage_analysis}
\label{sec:exp_errors}
Chained Stage 2 accuracy measures joint correctness of Stage 1 Anomaly Detection and Stage 2 Anomaly Recognition, where errors in either stage prevent downstream credit under the HIIP-Chain. After Stage 2 succeeds, Anomaly Localization, Anomaly Description, and Risk Analysis in Stage 3 are scored in parallel without mutual dependencies, with Anomaly Localization recording the lowest accuracy among T3--T5 for every baseline across both viewpoints in Table~\ref{tab:benchmark_results}. Stage 4 Maintenance Resolution requires correct answers in Stage 1, Stage 2, at least one Stage 3 question, and Stage 4 itself, reported as $\operatorname{Final} = \operatorname{Acc}_6$.

\paragraph{Category Coverage and Viewpoint Differences.}
\label{app:infrastructure_analysis}
\label{sec:exp_facilities}
The Ground and UAV test sets cover their complete taxonomies of 100 and 17 anomaly categories, while their training splits cover 66 and 15 categories, respectively. The 34 Ground categories absent from training and validation account for 112 test records, and 2 single-record UAV categories appear only in test, meaning these classes have no same-category training records in the reference banks. Appendix~\ref{app:category_inventory} lists the complete split-specific inventories. Overall scores pool the two test sets weighted by record count.

\begin{figure*}[!t]
\centering
\includegraphics[width=\linewidth,keepaspectratio,pagebox=mediabox,trim=8bp 157.5bp 8bp 158bp,clip]{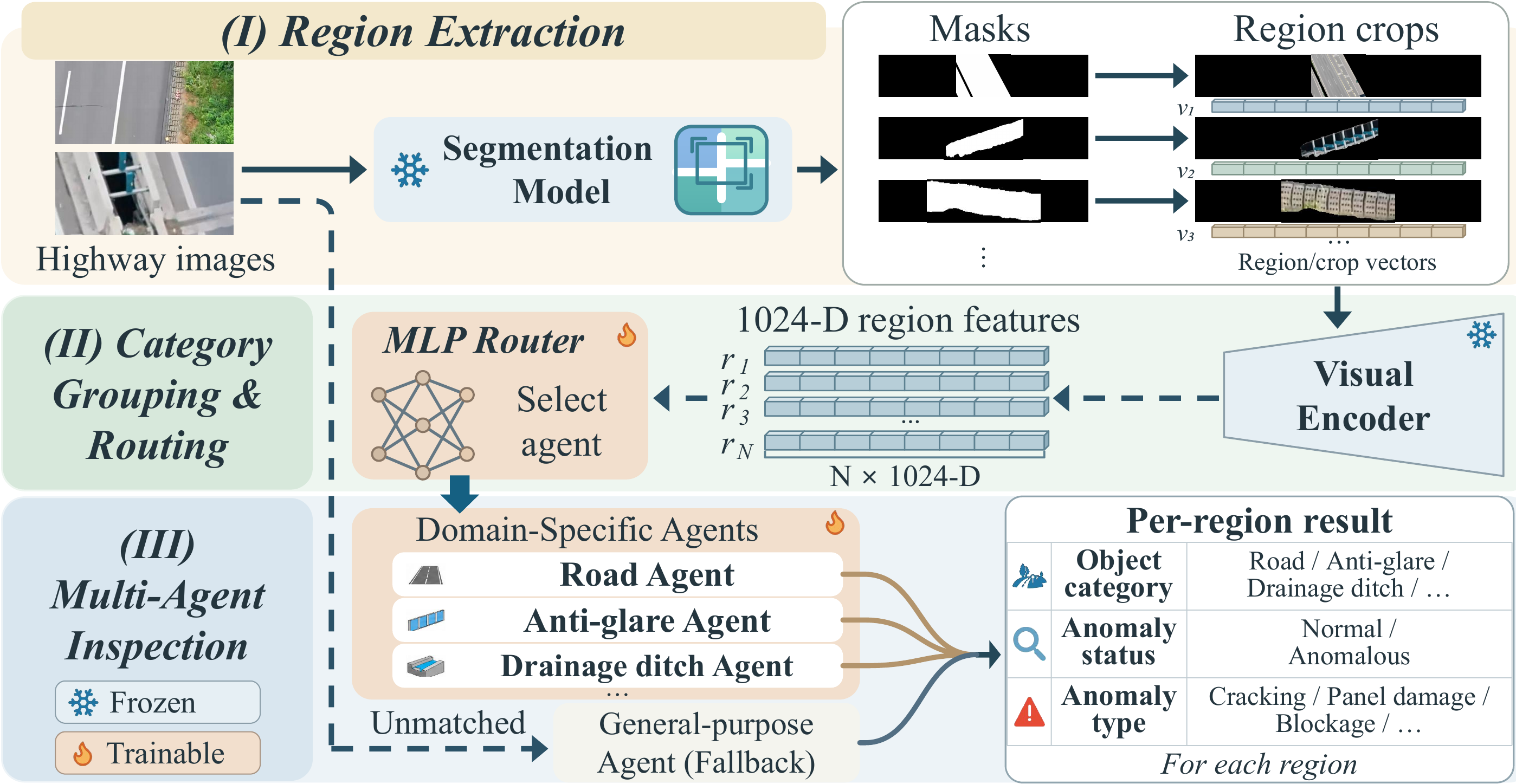}
\caption{\textbf{Trainable Segmentation-Guided Specialist Pipeline} for full-taxonomy T2 recognition, integrating SAM3 region isolation, trainable MLP feature routing, and QLoRA specialist diagnosis.}
\label{fig:highwayverse_pipeline}
\end{figure*}

\section{Extended Full-Taxonomy Recognition}
\label{app:full_taxonomy}
\label{sec:exp_full_taxonomy}

\paragraph{Evaluation setting.}
Table~\ref{tab:uav_full_taxonomy} compares independent T2 accuracy under four-option and full-taxonomy protocols on the same 963 UAV images. This supplementary roster comprises 9 open-source baselines, including Qwen3-VL-2B, plus GPT-4o and GPT-4o mini; it differs from the 12-model main comparison. Four-option baselines use one question per request. Both accuracy columns evaluate T2 over all 963 images regardless of T1 correctness.

\paragraph{Prompting and decoding.}
Four-option open-source results use the settings in Section~\ref{sec:exp_setup}; GPT-4o uses temperature 0.7, top-$p$ 0.8, seed 42, and a 12-token limit. The 17-class protocol supplies the complete category list and requests one category name, with a 32-token limit, greedy decoding for open-source models, and temperature 0 for GPT models. We remove at most one leading \texttt{-} list marker before exact matching the complete category name. The reported $\Delta$ therefore compares protocols that differ in prompt format, candidate space, and decoding settings.

\begin{table}[!t]
  \centering
  \captionsetup{position=top,font=small}
  \caption{Independent T2 accuracy on 963 UAV images, reported in percent. The 4-option scores use separate-question requests. $\Delta$: 17-class minus 4-option accuracy in percentage points. Bold rows denote the Segmentation-Guided Specialist Pipeline.}
  \label{tab:uav_full_taxonomy}
  \footnotesize
  \setlength{\tabcolsep}{2.5pt}
  \renewcommand{\arraystretch}{1.12}
  \definecolor{oursfulltax}{HTML}{E8F2FA}
  \begin{tabular*}{\columnwidth}{@{\extracolsep{\fill}}lccc@{}}
    \toprule
    & \textbf{4-option} & \textbf{17-class} & \\
    \cmidrule(lr){2-2}\cmidrule(lr){3-3}
    \textbf{Method} & \textbf{Acc.} & \textbf{Acc.} & $\boldsymbol{\Delta}$ \\
    \midrule
    \multicolumn{4}{@{}l}{\textit{Open-source models}\enspace($N=963$)} \\
    Qwen3.5-9B & 74.97 & 48.29 & $-26.69$ \\
    MiniCPM-V4.5 & 71.44 & 37.28 & $-34.16$ \\
    Qwen3-VL-8B & 72.48 & 44.55 & $-27.93$ \\
    Qwen3.5-27B & 84.22 & 48.39 & $-35.83$ \\
    InternVL3.5-8B & 66.56 & 32.92 & $-33.64$ \\
    Qwen2.5-VL-3B & 53.69 & 27.10 & $-26.58$ \\
    Qwen2.5-VL-7B & 62.93 & 40.91 & $-22.01$ \\
    Qwen3-VL-2B & 52.54 & 31.98 & $-20.56$ \\
    Qwen3-VL-32B & 81.41 & 46.00 & $-35.41$ \\
    \rowcolor{oursfulltax}
    \textbf{MiniCPM-V4.5 specialist} & \textbf{79.96} & \textbf{58.98} & $\boldsymbol{-20.98}$ \\
    \midrule
    \multicolumn{4}{@{}l}{\textit{Closed-source models}\enspace($N=963$)} \\
    GPT-4o & 83.07 & 49.84 & $-33.23$ \\
    GPT-4o mini & 71.86 & 49.43 & $-22.43$ \\
    \rowcolor{oursfulltax}
    \textbf{GPT-4o specialist} & \textbf{85.88} & \textbf{66.25} & $\boldsymbol{-19.63}$ \\
    \bottomrule
  \end{tabular*}
\end{table}

\paragraph{Full-taxonomy recognition remains difficult.}
All 11 baselines have lower full-taxonomy accuracy. The largest gap occurs for Qwen3.5-27B, whose independent T2 accuracy is 84.22\% with four options and 48.39\% under the 17-class protocol. GPT-4o mini drops from 71.86\% to 49.43\%, a decline of 22.43 percentage points. Recognition across the complete taxonomy remains challenging even for models with stronger four-option performance.

\subsection{Segmentation-Guided Specialist Pipeline}
\label{app:specialist_pipeline}
\label{sec:method_sam}
\label{sec:method_full_taxonomy}

This UAV-only supplementary experiment evaluates a separately trained \textbf{Segmentation-Guided Specialist Pipeline} for full-taxonomy T2 recognition, as illustrated in Figure~\ref{fig:highwayverse_pipeline}. Table~\ref{tab:uav_full_taxonomy} reports the specialist results separately from LaneRAG.

\paragraph{Region Extraction.}
Given input image $I$, frozen SAM3~\citesupp{Carion_2025_SAM3} extracts $N$ candidate infrastructure masks $\{M_n\}_{n=1}^{N}$ to generate background-suppressed region crops $\widetilde{I}_n = \operatorname{Crop}(I \odot M_n)$ alongside pooled regional embeddings $\mathbf{r}_n \in \mathbb{R}^{1024}$. 

\paragraph{Routing and Specialist Diagnosis.}
A lightweight multilayer-perceptron router $R_{\psi}: \mathbb{R}^{1024} \rightarrow \mathbb{R}^{K}$, trained with cross-entropy supervision on training-split crops, maps each $\mathbf{r}_n$ to its host infrastructure family $\hat{k}_n = \operatorname*{arg\,max}_{k} [R_{\psi}(\mathbf{r}_n)]_k$ and dispatches $\widetilde{I}_n$ to the corresponding domain-specific agent $\mathcal{A}_{\hat{k}_n}$, backed by an unmatched fallback agent for long-tail structures. Each specialist shares a frozen 4-bit quantized VLM backbone~\cite{Bai_2025_Qwen3VL} equipped with a trainable family-specific QLoRA adapter~\citesupp{Dettmers_2023_QLoRA}, outputting regional anomaly status $\hat{a}_n \in \{\text{normal}, \text{anomalous}\}$ and full-taxonomy defect category $\hat{y}_n$. 

\paragraph{Prediction Aggregation.}
Regional expert categories are mapped to the answer space of each evaluation protocol, with a general-model fallback when no expert reports an anomaly. For Table~\ref{tab:uav_full_taxonomy}, the following paragraphs specify the composition and conflict handling.

\paragraph{Composition with single-question baselines.}
The two variants share the same regional expert predictions: 338 images have a unique expert category, 7 have conflicting expert categories, and 618 use the general-model fallback. Four-option scores retain the expert decisions and use the corresponding single-question T2 answers on fallback images. The 17-class scores retain the independent full-taxonomy predictions. Expert categories are mapped to the candidate options in the four-option setting; unresolved category conflicts count as errors in the 17-class setting.

\paragraph{Specialist collaboration improves recognition.}
With single-question fallback, the MiniCPM-V4.5 specialist reaches 79.96\% four-option accuracy, compared with its 71.44\% baseline; the GPT-4o variant reaches 85.88\%, compared with 83.07\%. Their 17-class accuracies remain 58.98\% and 66.25\%, exceeding their respective baselines by 21.70 and 16.41 percentage points. The corresponding differences between 17-class and four-option accuracy are $-20.98$ and $-19.63$ percentage points.

\makeatletter
\if@firstcolumn\newpage\fi
\makeatother
\section{Limitations and Interpretation}
\label{app:limitations}

\paragraph{Dataset and Evaluation Scope.}
Ground and UAV differ in both viewpoint and category composition. The main benchmark uses multiple-choice questions, leaving open-ended inspection reporting and its evaluation for future work. Four-option and full-taxonomy evaluation differ in prompt format, candidate space, output budget, and decoding, so neither comparison isolates a single factor. The test sets cover all 100 Ground and 17 UAV categories. As detailed in Appendix~\ref{app:category_inventory}, sparse test coverage can limit the stability of category-level results.

\paragraph{Attribution.}
Specialist results combine region extraction, routing, and supervised adaptation; their individual contributions require separate ablations. Category frequencies do not measure model pretraining exposure, and answer co-occurrence does not identify internal reasoning cues.

\end{document}